\documentclass[journal]{IEEEtran}
\usepackage{graphicx}
\usepackage{float}
\usepackage{subfigure}
\usepackage{epsfig}
\usepackage{epstopdf}
\usepackage{indentfirst}
\usepackage{amsfonts}
\usepackage{amsmath,amssymb}
\usepackage{algorithm}
\usepackage{algorithmic}
\usepackage{verbatim}
\usepackage{multirow}
\usepackage{array}
\usepackage{color}
\usepackage{cite}
\usepackage{makecell}
\usepackage{balance}
\usepackage{hyperref}
\usepackage{balance}

\ifCLASSINFOpdf

\else

\fi

\begin{document}

\title{Maximum Entropy Subspace Clustering Network}
\author{
    Zhihao~Peng, ~\IEEEmembership{Graduate Student Member,~IEEE,}
	Yuheng~Jia,
 	Hui~Liu,
	Junhui~Hou, ~\IEEEmembership{Senior Member,~IEEE}, \\ and
  	Qingfu~Zhang, ~\IEEEmembership{Fellow,~IEEE}
\thanks{This work was supported by the Hong Kong RGC under Grants 9048123 (CityU 21211518), 9042820 (CityU 11219019), and 9042955 (CityU 11202320).}	
\thanks{Z. Peng, H. Liu, J. Hou and Q. Zhang are with the Department of Computer Science, City University of Hong Kong, Kowloon, Hong Kong 999077 (e-mail: zhihapeng3-c@my.cityu.edu.hk; hliu99-c@my.cityu.edu.hk; jh.hou@cityu.edu.hk; qingfu.zhang@cityu.edu.hk)}
\thanks{Y. Jia is with the School of Computer Science and Engineering, Southeast University, and also with Key Laboratory of Computer Network and Information Integration (Southeast University), Ministry of Education, Nanjing, 211189, China (e-mail: yhjia@seu.edu.cn).}
\thanks{Corresponding author: J. Hou (e-mail: jh.hou@cityu.edu.hk)}
}

\markboth{Revised Manuscript submitted IEEE Transactions on Circuits and Systems for Video Technology}
{Shell \MakeLowercase{\textit{et al.}}: Bare Demo of IEEEtran.cls for IEEE Journals}

\maketitle

\begin{abstract}
Deep subspace clustering networks have attracted much attention in subspace clustering, in which an auto-encoder non-linearly maps the input data into a latent space, and a fully connected layer named self-expressiveness module is introduced to learn the affinity matrix via a typical regularization term (e.g., sparse or low-rank). 
However, the adopted regularization terms ignore the connectivity within each subspace, limiting their clustering performance. In addition, the adopted framework suffers from the coupling issue between the auto-encoder module and the self-expressiveness module, making the network training non-trivial. To tackle these two issues, we propose a novel deep subspace clustering method named Maximum Entropy Subspace Clustering Network (MESC-Net). 
Specifically, MESC-Net maximizes the entropy of the affinity matrix to promote the connectivity within each subspace, in which its elements corresponding to the same subspace are uniformly and densely distributed. Meanwhile, we design a novel framework to explicitly decouple the auto-encoder module and the self-expressiveness module. Besides, we also theoretically prove that the learned affinity matrix satisfies the block-diagonal property under the assumption of independent subspaces. 
Extensive quantitative and qualitative results on commonly used benchmark datasets validate MESC-Net significantly outperforms state-of-the-art methods. The code is publicly available at \url{https://github.com/ZhihaoPENG-CityU/MESC}.
\end{abstract}


\IEEEpeerreviewmaketitle

\section{Introduction}

\IEEEPARstart{C}{lustering} aims to partition a collection of samples into different groups such that samples in the same group are similar and samples from different groups are dissimilar. Many real-world applications can be categorized as a clustering problem, e.g., signal propagation \cite{liu2019imbalance,jia2020pairwise,jia2020constrained}, object clustering \cite{yang2019subspace,xu2020autoencoder,jia2020semisupervised,9336710}, and transfer clustering \cite{peng2019active,peng2020non}. 
Recently, deep learning-based clustering methods have demonstrated strong competitiveness, in which the deep self-expressiveness-based subspace clustering \cite{peng2016deep,ji2017deep,chen2018subspace}, as a kind of popular deep learning-based clustering method, is benefited from the assumption that each data sample can be represented as a linear combination of other samples in the same subspace. Specifically, deep self-expressiveness-based subspace clustering methods can non-linearly map input data into a latent space and learn an affinity matrix with a typical regularization (e.g., sparse or low-rank) simultaneously.
\begin{figure}
	\centering
	\includegraphics [width=0.98\columnwidth]{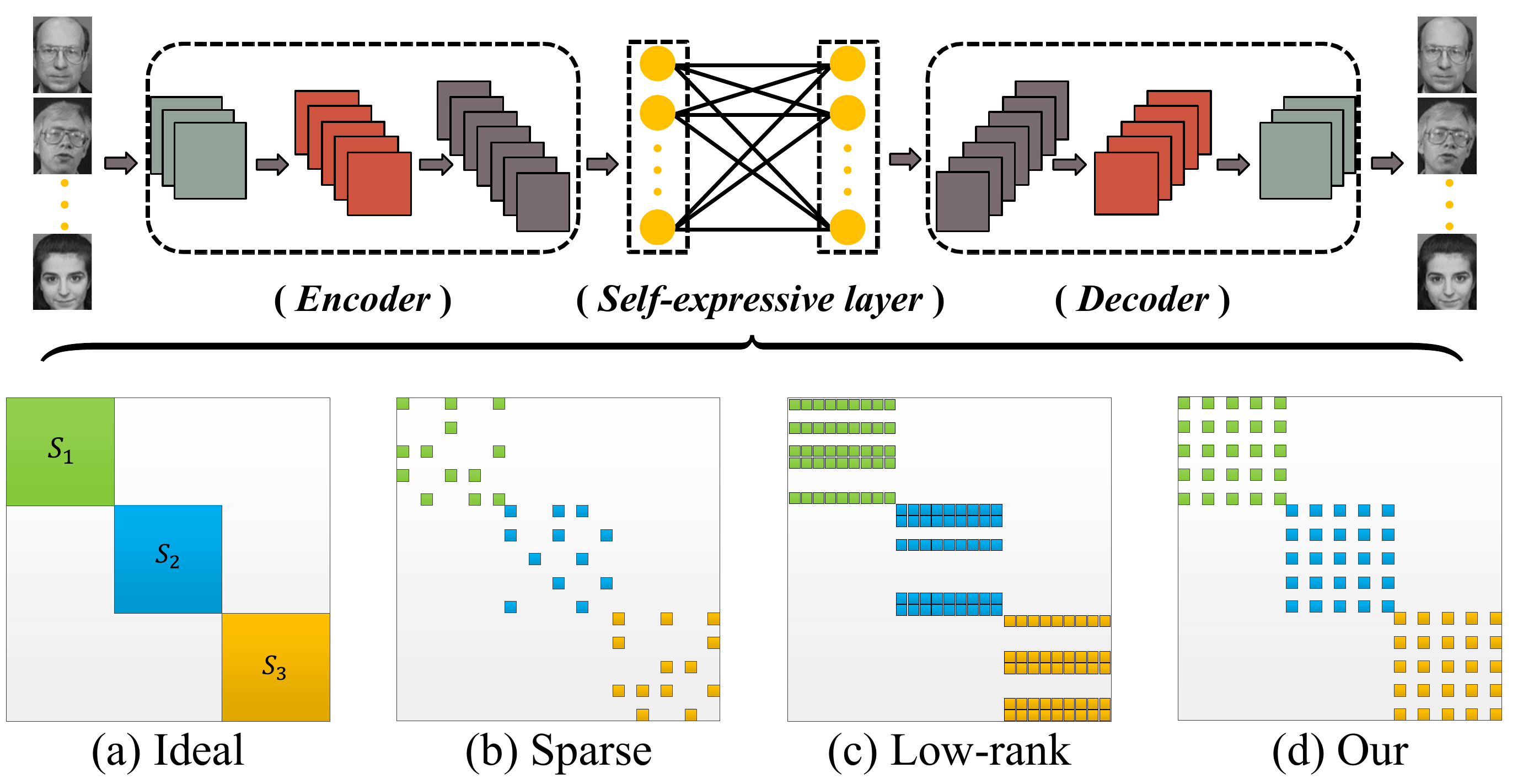}
	\caption{\textbf{The upper part} is the generally adopted architecture of deep self-expressiveness-based subspace clustering network (DSC-Net \cite{ji2017deep}), which is an auto-encoder (i.e., a series of convolutional layers) with a self-expressive layer (i.e., a fully connected layer without any activation and bias) inserted between the encoder and the decoder. Input data are first mapped into a latent space by the encoder, and an affinity matrix is then learned by self-expressing the resulting features, which is driven by the reconstruction of input data from the features in the decoder. At the same time, a typical regularization like sparse or low-rank is imposed on the affinity matrix. \textbf{The lower part} is the schematic diagrams of the learned affinity matrices under various regularization techniques. (a) the ideal affinity matrix structure for input data, which are assumed to be distributed in three independent subspaces $S_1, S_2$ and $ S_3$; (b) Learned affinity matrix by the sparse regularization; (c) Learned affinity matrix by the low-rank regularization; (d) Ours.}
	\label{fig:overview}
\end{figure}
    For example, Peng \emph{et al.} \cite{peng2016deep} proposed to incorporate the sparsity structure into the hidden representation learning. Chen \emph{et al.} \cite{chen2018subspace} combined the auto-encoder and low-rank regularization to find the underlying lowest rank representation. Ji \emph{et al.} \cite{ji2017deep} proposed a popular deep subspace clustering architecture with sparse and Tikhonov regularization (i.e., DSC-Net-L1 and DSC-Net-L2) by introducing a fully connected layer between the encoder and the decoder to simulate the self-expressiveness procedure, as illustrated in the upper part of Figure \ref{fig:overview}. Besides, various DSC-Net variants have been proposed \cite{lei2020deep,kang2020structure,gao2020cross}. 

\begin{figure}
	\centering
	\includegraphics [width=0.98\columnwidth]{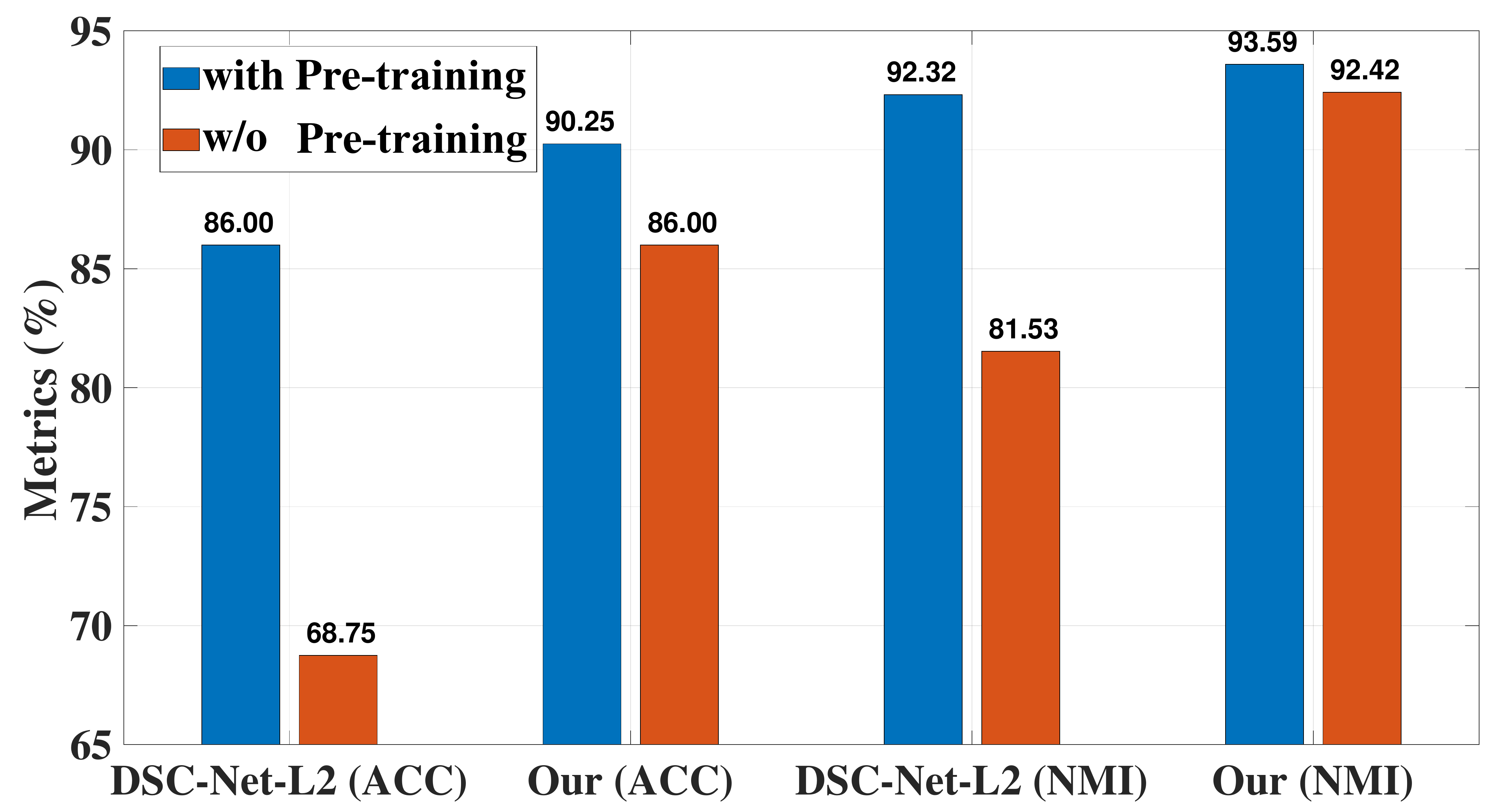}
	\caption{Illustration of the sensitivity of DSC-Net to the pre-training process. The experiment was conducted on the ORL dataset. Here, we use accuracy (ACC) and normalized mutual information (NMI) as the metrics. Compared with DSC-Net-L2, our method is less sensitive to the pre-training process, which is credited to the proposed decoupling network.}
	\label{fig:mtv2}
\end{figure}

    However, we observe that: (1) the adopted regularization terms by the above-mentioned methods ignore the connectivity within each subspace, which compromises the subsequent spectral clustering \cite{ng2002spectral} to some extent. In particular, the connectivity within each subspace (i.e., the relations between data) appears to be particularly important in the case that most real-world data do not satisfy the independent subspace assumption; and (2) in the DSC-Net-based frameworks, the auto-encoder module and the self-expressiveness module are tightly coupled, making the network training non-trivial, i.e., the clustering performance of directly trained DSC-Net will drop significantly (see Figure \ref{fig:mtv2}). To this end, Ji \emph{et al.} \cite{ji2017deep} designed the pre-training and fine-tuning strategy. However, such a strategy consumes additional computing time and resources.
    
    To remedy these two issues, we propose a novel deep self-expressiveness-based subspace clustering method, namely Maximum Entropy Subspace Clustering Network (MESC-Net). Specifically, we use the Maximum Entropy (ME) regularization to constrain the learning of the affinity matrix in which its elements corresponding to the same subspace are uniformly and densely distributed. The lower part of Figure \ref{fig:overview} visually demonstrates the advantage of the ME prior over the traditionally used sparse and low-rank priors. Furthermore, we theoretically prove that the learned affinity matrix satisfies the block-diagonal property \cite{lu2018subspace} under the independent subspaces. Moreover, we explicitly decouple the auto-encoder module and the self-expressiveness module. The extensive quantitative and qualitative comparisons conducted on one toy dataset and seven commonly used benchmark datasets validate the superiority of MESC-Net over state-of-the-art methods.

    {The rest of this paper is organized as follows. We review some related works in Section \uppercase\expandafter{\romannumeral2}. In Section \uppercase\expandafter{\romannumeral3}, we introduce our method and provide the visual illustration and the block-diagonal property analysis, followed by the experimental results and analyses in Section \uppercase\expandafter{\romannumeral4}. Finally, we conclude the paper in Section \uppercase\expandafter{\romannumeral5}.} 

    \section{Related Work}

	Throughout this paper, scalars are denoted by italic lower case letters, vectors by bold lower case letters, matrices by upper case ones, and operators by calligraphy ones, respectively. Given a matrix $\mathbf{A}\in\mathbb{R}^{m\times n}$, $\left\|\mathbf{A}\right\|_0, \left\|\mathbf{A}\right\|_1, \left\|\mathbf{A}\right\|_F$ and $\left\|\mathbf{A}\right\|_*$ denote its $\ell_0$ norm (i.e., $\left\|\mathbf{A}\right\|_0=\mathsf{Card}\left\{i,j:a_{i,j}\neq0 \right\}$ with $\mathsf{Card}\{\cdot\}$ computing the cardinality of a finite set), $\ell_1$ norm (i.e., $\left\|\mathbf{A}\right\|_1= \sum_{i=1}^{m}\sum_{j=1}^{n}{\left| a_{i,j} \right|}$), Frobenius norm (i.e., $\left\|\mathbf{A}\right\|_F=\sqrt{\sum_{i=1}^{m}\sum_{j=1}^{n}{\left| a_{i,j} \right|}^2}$), and nuclear norm (i.e., $\left\|\mathbf{A}\right\|_*=\sum_{i=1}^{\min\{m,n\}}{\sigma_{i}\left(\mathbf{A}\right)}$ with $\sigma_i(\mathbf{A})$ being the $i$-th singular value of $\mathbf{A}$), respectively \cite{zhang2017matrix}. Table \ref{tab:notation} summarizes the main notations used throughout the paper. 

    \begin{table}[t]
        \caption{Main notations and descriptions.}
        \label{tab:notation}
        \centering
            \begin{tabular}{l|c}
            \hline\hline
            Notations                   & Descriptions                                  \\ \hline
            $\mathbf{X}$                & Input data                                \\
            $ d_1\times d_2\times1 $    & The dimension of one input sample             \\
            $n$                         & The number of samples                         \\
            $d$                         & The dimension of latent features           \\
            $\Theta_E$                  & The encoder parameters                        \\
            $f_E(\cdot ; \cdot)$        & The nonlinear mapping of the encoder          \\
            $\hat{\mathbf{Z}}$          & Output of the encoder                     \\
            $p_1\times p_2\times p_3$   & The dimension of $\hat{\mathbf{Z}}$           \\
            $\Theta_D$                  & The decoder parameters                        \\
            $f_D(\cdot ; \cdot)$        & The nonlinear mapping of the decoder          \\
            $\mathbf{Z}$                & The reshaped features via $\hat{\mathbf{Z}}$  \\
            $p$                         & The dimension of $\mathbf{Z}$                 \\
            $\mathbf{C}$                & The learned affinity matrix                   \\ 
            $f_C(\cdot ; \cdot)$        & \makecell[l]{The reshape operation of input data and \\ the mapping of the self-expressiveness module}\\ \hline    \hline
            \end{tabular}
    \end{table}

    \subsection{Traditional Self-expressiveness-based Methods}

	Let $\mathbf{Z}\in\mathbb{R}^{d\times n}$ denote the representation matrix with $d$ and $n$ being the dimension of features and number of samples, respectively. The self-expressiveness-based models can be generally formulated as
	\begin{equation}
	\begin{aligned}
	\min_{\mathbf{C}} f(\mathbf{C}) \quad \rm{s.t.}\quad \mathbf{Z} = \mathbf{Z}\mathbf{C},
	\label{eq:SC}
	\end{aligned}
	\end{equation}
	where $\mathbf{C}\in\mathbb{R}^{n\times n}$ is the affinity matrix, and $f(\mathbf{C})$ is the implicit regularization on $\mathbf{C}$ to promote a unique and meaningful solution under the assumption of self-expressiveness. According to the imposed regularization on $\mathbf{C}$, the existing approaches can be mainly divided into two categories: the sparse subspace clustering (SSC) based methods \cite{elhamifar2013sparse,patel2014kernel,wang2017subspace,chen2018structured,liu2020locality} and the low-rank representation (LRR) based methods \cite{liu2012robust,xiao2015robust,xing2020robust,shen2021fast,chen2021low}. To be specific, SSC \cite{elhamifar2013sparse} was proposed to find the nontrivial sparse representation, of which the $\ell_1$ norm minimization (i.e., the tightest convex relaxation of the $\ell_0$ norm) is employed. In addition, Wang \emph{et al.} \cite{wang2017subspace} explored the relations between data by integrating SSC and the least squares regression to improve the segmentation performance. Chen \emph{et al.} \cite{chen2018structured} coupled the self-representation matrix and the segmentation matrix to encourage the within-cluster grouping. Since SSC based methods can only capture the linearly-relationship among samples, Kernel Sparse Subspace Clustering ({KSSC}) \cite{patel2014kernel} was proposed to exploit the nonlinear structure information in input space. However, in KSSC, the kernel function and the associated hyper-parameters are challenging to determine \cite{kang2017twin}. LRR \cite{liu2012robust} and Robust Kernel Low-Rank Representation ({RKLRR}) \cite{xiao2015robust} are the other two popular clustering methods that capture the global structure of samples with a low-rank constraint for coping with the linear and nonlinear data, respectively. Like SSC and KSSC, LRR and RKLRR also suffer from the above-mentioned issues.
	
	\subsection{Deep Self-expressiveness-based Methods}
	Due to the breakthroughs in deep learning, numerous deep subspace clustering methods \cite{ji2017deep,chen2018subspace,kang2020structure} were proposed to learn a nonlinear mapping of the data that is well-adapted to clustering. For example, Ji \emph{et al.} \cite{ji2017deep} proposed the deep subspace clustering network (i.e., DSC-Net) to achieve the nonlinear mapping by an auto-encoder and mimic the self-expressiveness with an $\ell_1$ regularization (i.e., DSC-Net-L1) or an $\ell_2$ regularization (i.e., DSC-Net-L2). Specifically, let $\mathbf{X} = [\mathbf{x}_1,...,\mathbf{x}_{n}]^\mathsf{T}$ denote input data with $n$ being the number of samples, $\mathbf{X}_r$ denote the reconstructed data by the auto-encoder, $\Theta_E$ and $\Theta_D$ denote the encoder and decoder parameters, respectively, and $\hat{\mathbf{Z}}$ denote the extracted latent representation. A fully connected layer (without any activation and bias) is introduced between the encoder and the decoder to learn an affinity matrix $\mathbf{C}$ with a typical regularization term. The loss function of DSC-Net is expressed as

	\begin{equation}
	\begin{aligned}
	&\!\!\min_{\mathbf{C}}\frac{1}{2} \left\| \mathbf{X} \!-\! \mathbf{X}_r \right\|^2_F \!+\!\!\left\| \mathbf{C} \right\|_p \!+\!\!\left\| \mathbf{Z} \!-\! \mathbf{Z}\mathbf{C} \right\|^2_F \\
	&\rm{s.t.} \quad diag(\mathbf{C})=\mathbf{0},
	\label{eq:SE}
	\end{aligned}
	\end{equation}
	where $\hat{\mathbf{Z}}=f_E(\mathbf{X};\Theta_E)$ is reshaped as a data matrix $\mathbf{Z}$, $\mathbf{X}_r=f_D(f_C(f_E(\mathbf{X};\Theta_E);\mathbf{C});\Theta_D)$, minimizing $\left\| \mathbf{Z} - \mathbf{Z}\mathbf{C} \right\|^2_F$ aims to achieve self-expressiveness, $\left\| \mathbf{C} \right\|_p$ represents a typical matrix norm (e.g., $\|\mathbf{C}\|_1$ and $\|\mathbf{C}\|_F^2$ in \cite{ji2017deep}) to regularize $\mathbf{C}$, and $diag(\mathbf{C})=\mathbf{0}$ is used to prevent the trivial solution, i.e., $\mathbf{C}=\mathbf{I}$ with $\mathbf{I}$ being the identity matrix. Moreover, Chen \emph{et al.} \cite{chen2018subspace} proposed the low-rank constrained auto-encoder (LRAE) that combines LRR and the auto-encoder together to promote the non-linear modeling capability of LRR. Xue \emph{et al.} \cite{xue2019deep} integrated the deep matrix factorization, low-rank subspace learning, and multiple subspace ensemble in a unified framework. Zhu \emph{et al.} \cite{zhu2020sparse} introduced the sparse and low-rank constraints on the deep feature and the self-expressive matrix, respectively. Furthermore, based on the DSC-Net framework, Zhou \emph{et al.} \cite{zhou2019latent} learned a distribution-preserving latent representation by a distribution consistency loss. Kang \emph{et al.} \cite{kang2020structure} preserved the pairwise similarities between the data points by similarity preserving term. 
	
	{The methods mentioned above have improved the performance of traditional clustering methods to a large extent. However, as shown in Figure \ref{fig:overview}, the learned affinity matrices by both sparse and low-rank regularizations are incapable of being an ideal affinity matrix as they ignore the connectivity within each subspace. Besides, DSC-Net claimed that it is difficult to directly train the network with millions of parameters from scratch \cite{ji2017deep}. We reason that the DSC-Net-based framework suffers from the coupling issue, see Section \uppercase\expandafter{\romannumeral2}-D.}

	\subsection{Maximum Entropy}
    In information theory, entropy is a measure of the amount of information required on average to describe the random variable. Specifically, given a discrete random variable $\mathbf{X}$, its entropy $H(\mathbf{X})$ is defined as $H(\mathbf{X})=-\sum_{x \in \mathbf{X}} p(x) \log p(x)$, where $p(x)$ denotes the probability mass function of $\mathbf{X}$, and $0\log0$ is taken to be 0 \cite{cover1991elements}. Entropy reaches its maximum if and only if the distribution is uniform. In the past decades, various methods based on the maximum entropy principle have been proposed for clustering. For example, Krause \emph{et al.} \cite{krause2010discriminative} proposed to simultaneously partition the data and train a discriminative classifier, which achieves the class balance via maximum entropy. Aldana \emph{et al.} \cite{aldana2015clustering} explored the space of all possible probability distributions of the data to find one that maximizes entropy subject to extra conditions based on prior information about the clusters. Kalofolias \emph{et al.} \cite{kalofolias2016learn} and Bai \emph{et al.} \cite{icml2020_1982} proposed to use the maximum entropy to construct a Gaussian kernel graph, where the weight of the learned graph is equivalent to the weight of the RBF kernel graph.

    \subsection{Coupling Issue}
    In software engineering, the coupling measures the degree of interdependence between the involved modules, and a suitable framework should have the property of low coupling. Particularly, content coupling refers to that one module can modify another module's data, or control flow is passed from one module to the other module, which is the worst form among various couplings. 
    {
    DSC-Net suffers from the content coupling issue between the auto-encoder module and the self-expressiveness module, as the self-expressiveness module can modify the data of the auto-encoder module. To solve the resulting inconvenience, Ji \emph {et al.} \cite{ji2017deep} designed the pre-training and fine-tuning strategy, but it will consume additional computing time and resources. In the previous works, Seo \emph {et al.} \cite{seo2019deep} proposed a variant of DSC-Net, which can be regarded as achieving decoupling for the self-expressive matrix. However, it still suffers from the content coupling issue as the reconstruction feature is affected by the self-expressiveness module in the auto-encoder structure.}

	\section{Proposed Method}

	\subsection{Motivation and Overview}
	{As aforementioned, in DSC-Net and its variants, the adopted regularization terms on the affinity matrix (e.g., sparse and low-rank regularizations) ignore the connectivity within each subspace, limiting the subsequent spectral clustering performance. In addition, the DSC-Net-based frameworks suffer from the coupling issue, as the auto-encoder module and the self-expressiveness module are tightly coupled, making the training of the DSC-Net non-trivial.} 
    To this end, we propose a Maximum Entropy Subspace Clustering Network (MESC-Net) for clustering. Specifically, MESC-Net imposes the Maximum Entropy (ME) regularization on the affinity matrix to encourage the elements corresponding to the same subspace to be uniformly and densely distributed, benefiting the subsequent spectral clustering. Meanwhile, we design a novel decoupling framework for network training by explicitly separating the auto-encoder module and the self-expressiveness module. In addition, we also theoretically prove that the learned affinity matrix satisfies the block-diagonal property under the assumption of independent subspaces.

    \begin{figure}
		\centering
		\includegraphics [width=0.98\columnwidth]{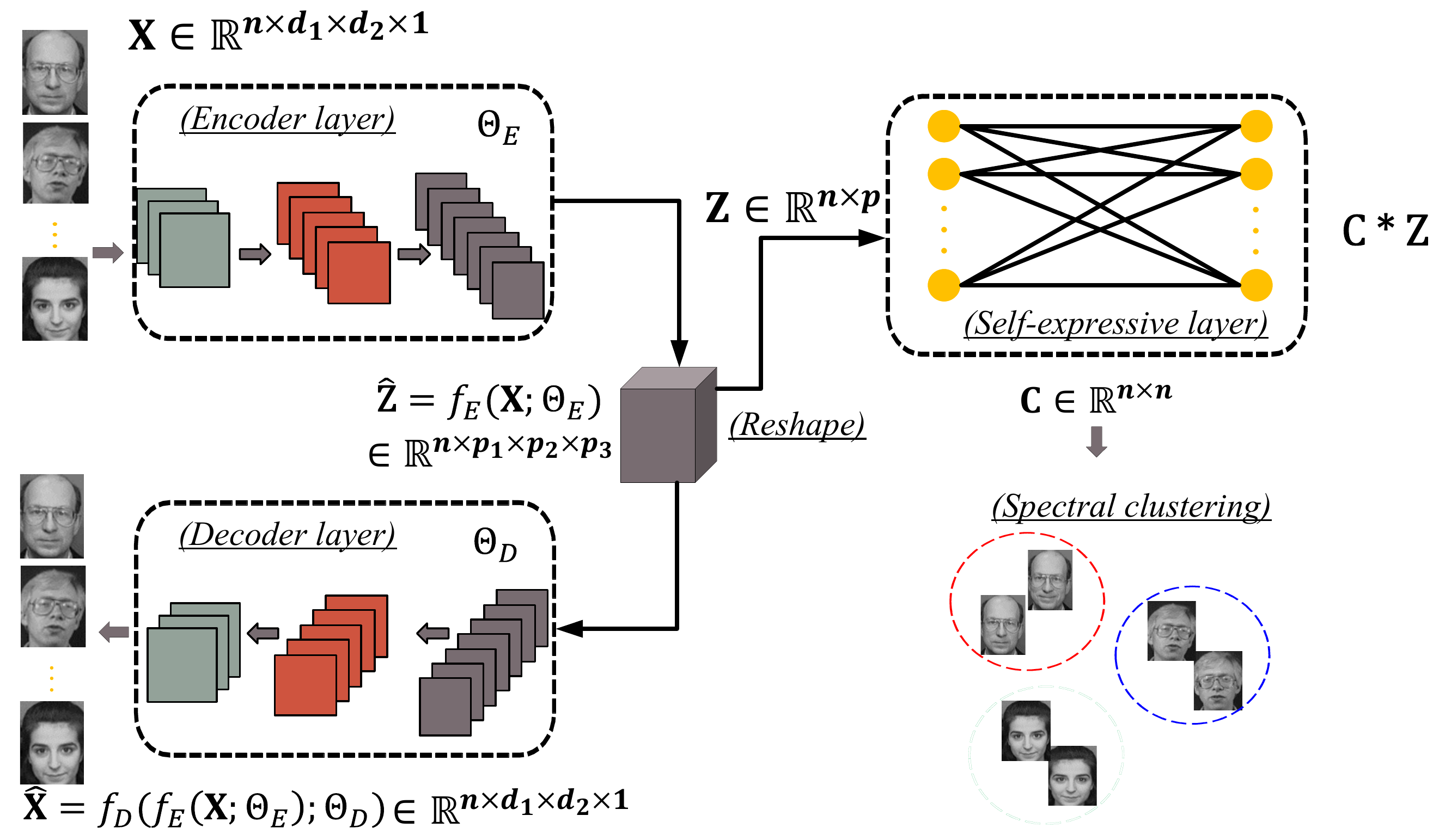}
		\caption{Illustration of the flowchart of the proposed MESC-Net. It consists of an auto-encoder module with a series of convolutional layers and a self-expressiveness module with a fully connected layer without any activation and bias. The auto-encoder module non-linearly maps input data into a latent space, and the self-expressiveness module learns an affinity matrix via a fully connected layer. Different from DSC-net, we explicitly decouple the auto-encoder module and the self-expressiveness module to avoid the coupling issue. More details of the notations can be found in Table \ref{tab:notation}.
		}
		\label{fig:framework}
\end{figure}

    \subsection{Model Formulation}
    The element of the affinity matrix $\mathbf{C}$ denoted by $c_{i,j}$ could be thought of as the similarity degree (i.e., probabilistic similarity) between the data samples $i$ and $j$. By applying the ME regularization on the self-expressiveness coefficients, as well as the fact $\max H(\mathbf{C}) = \min -H(\mathbf{C})$, the loss function for driving the learning of the affinity matrix can be written as
	\begin{equation}
	\begin{aligned}
	&\!\!\min_{\mathbf{C}} f(\mathbf{C}) =
	 -H(\mathbf{C}) = \sum_{i=1}^{n}\sum_{j=1}^{n} c_{i,j} \ln c_{i,j} \\ 
	&\rm{s.t.} \quad \mathbf{Z}=\mathbf{Z}\mathbf{C}, \quad \emph{c}_{\emph{i},\emph{j}} \geq 0, 
	\label{eq:Entropy}
	\end{aligned}
	\end{equation}
	where in the case of $c_{i,j} = 0$, the value of the corresponding summand $c_{i,j}\ln c_{i,j}$ is equal to 0 \cite{cover1991elements}. Intuitively, when a sample is represented only with the samples from the same subspace, minimizing Eq. (\ref{eq:Entropy}) will force the connections between samples belonging to the same subspace equally strong. In light of this, the previous constraint $diag(\mathbf{C})=\mathbf{0}$ is not needed.
	
    {Based on Eq. (\ref{eq:Entropy}), we propose MESC-Net, as shown in Figure \ref{fig:framework}.} Specifically, the auto-encoder module and the self-expressiveness module are separated at the network structure level, resulting in a reduction in the coupling issue. The overall loss function is written as

	\begin{equation}
	\begin{aligned}
	& \!\!\min_{\mathbf{C}} 
	\frac{1}{2} \left\| \mathbf{X} \!-\! \mathbf{X}_r \right\|^2_F 
	\!+\! \lambda_1\!\sum_{i=1}^{n}\sum_{j=1}^{n} c_{i,j} \!\ln c_{i,j}  \!+\!\lambda_2\!\left\| \mathbf{Z} \!- \!\mathbf{Z}\mathbf{C}  \right\|^2_F \\
	& \rm{s.t.} \quad c_{i,j} \geq 0,
	\label{eq:MESC-Net}
	\end{aligned}
	\end{equation}
	where $\mathbf{X}_r=f_D(f_E(\mathbf{X};\Theta_E);\Theta_D)$, $\lambda_1>0$ and $\lambda_2>0$ are the trade-off parameters. 

    \subsection{Visual Illustration of the Learned Affinity Matrix}
    
\begin{figure}
		\centering
		\includegraphics[width=0.98\columnwidth]{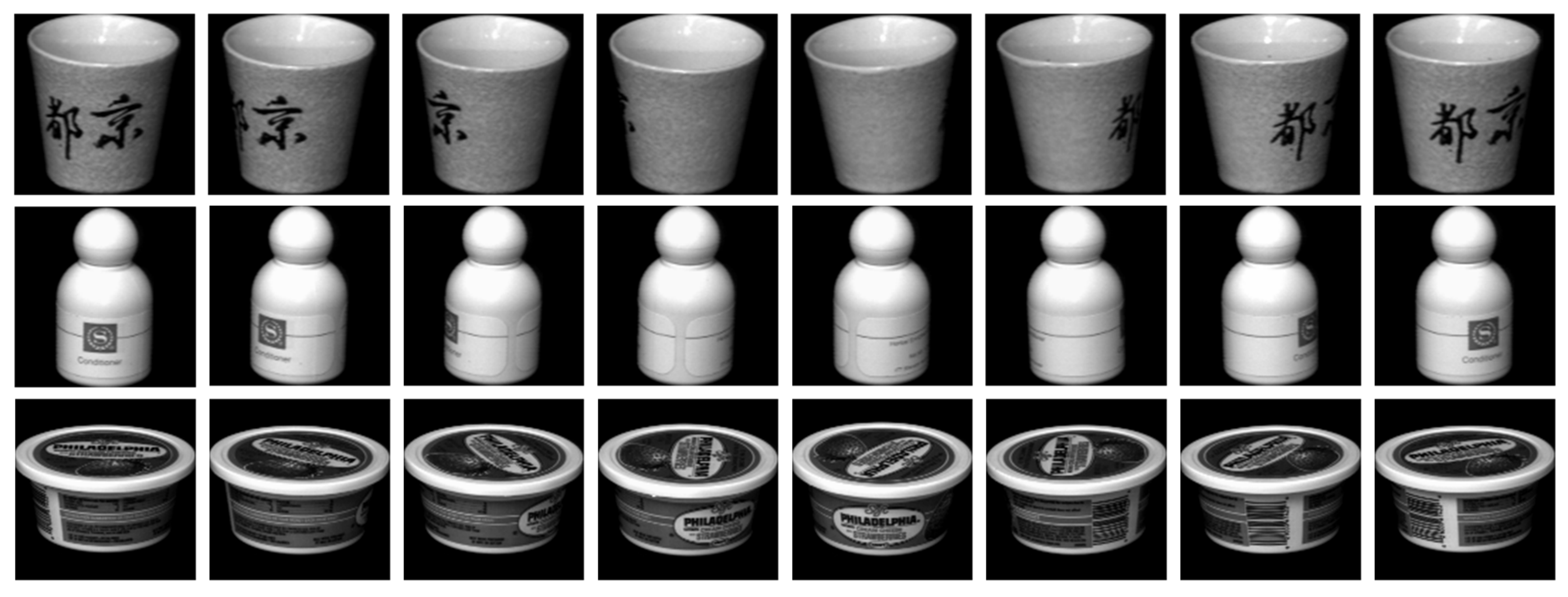}
		\caption{Sample images from the toy dataset.}
		\label{fig:Image}
	\end{figure}

\begin{figure*}
		\centering
		\subfigure[Ideal]{
			\includegraphics [width=0.36\columnwidth]{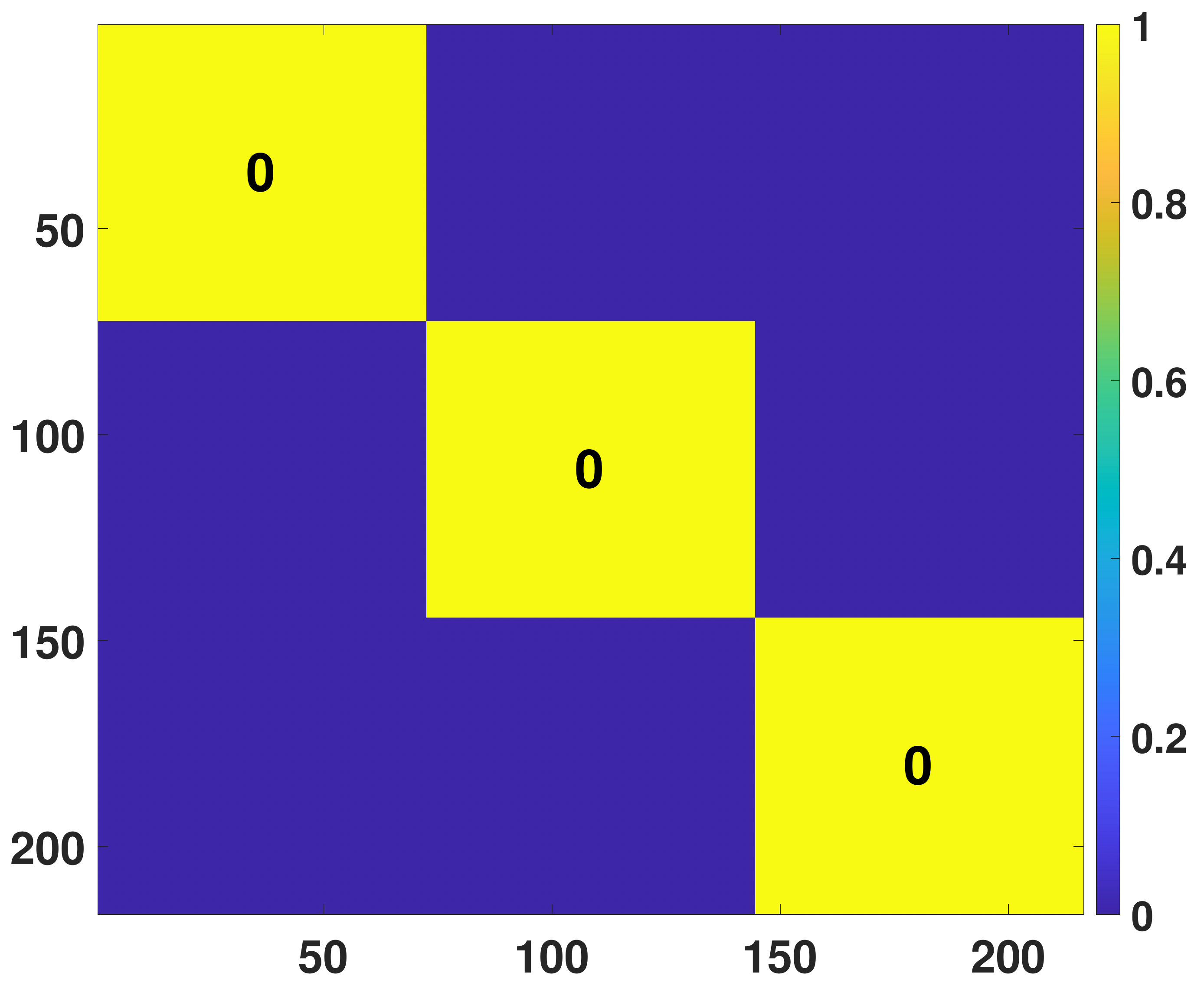}
		}
		\subfigure[$\|\mathbf{C}\|_1$]{
			\includegraphics [width=0.36\columnwidth]{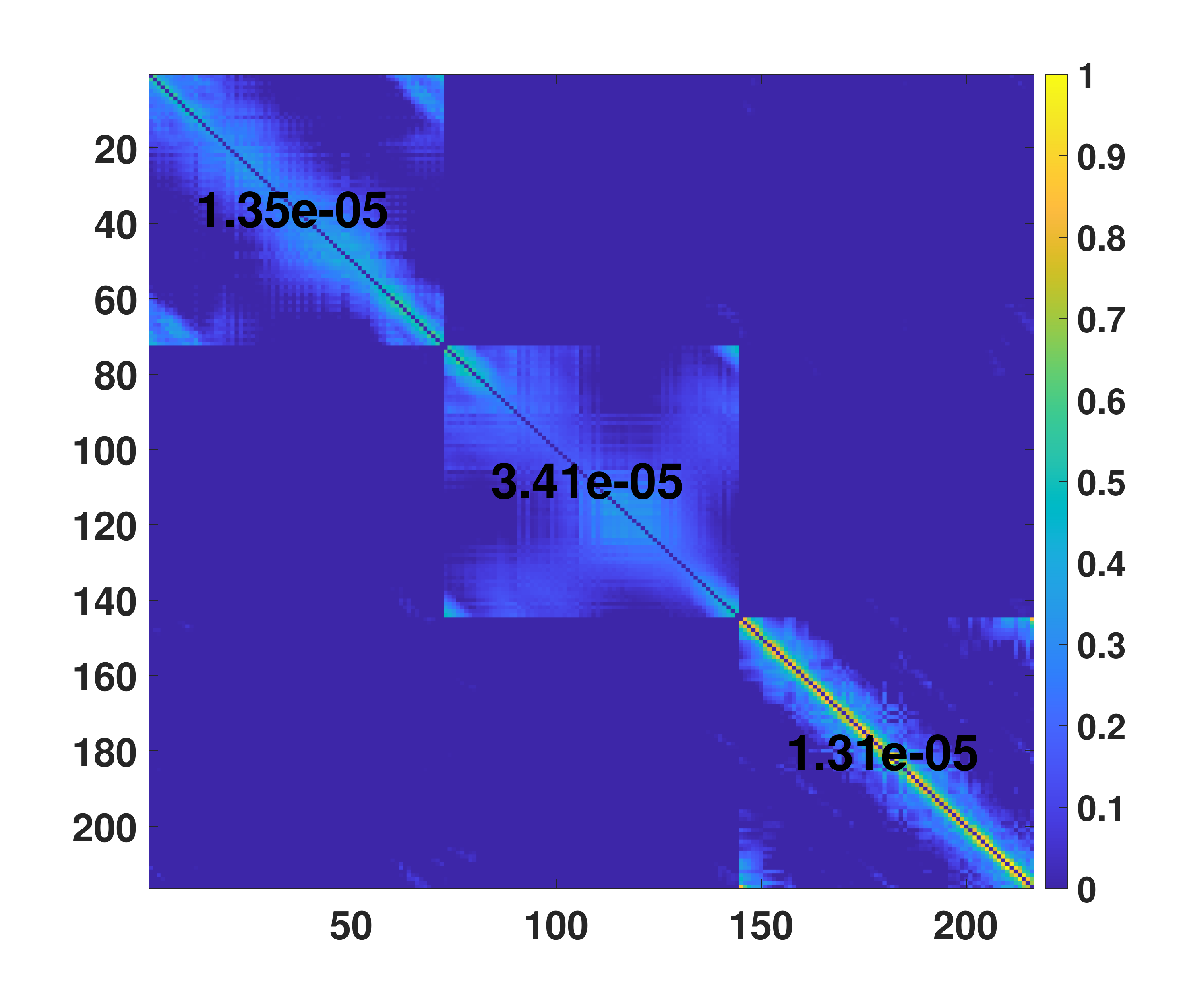}
		}
		\subfigure[$\|\mathbf{C}\|_F^2$]{
			\includegraphics [width=0.36\columnwidth]{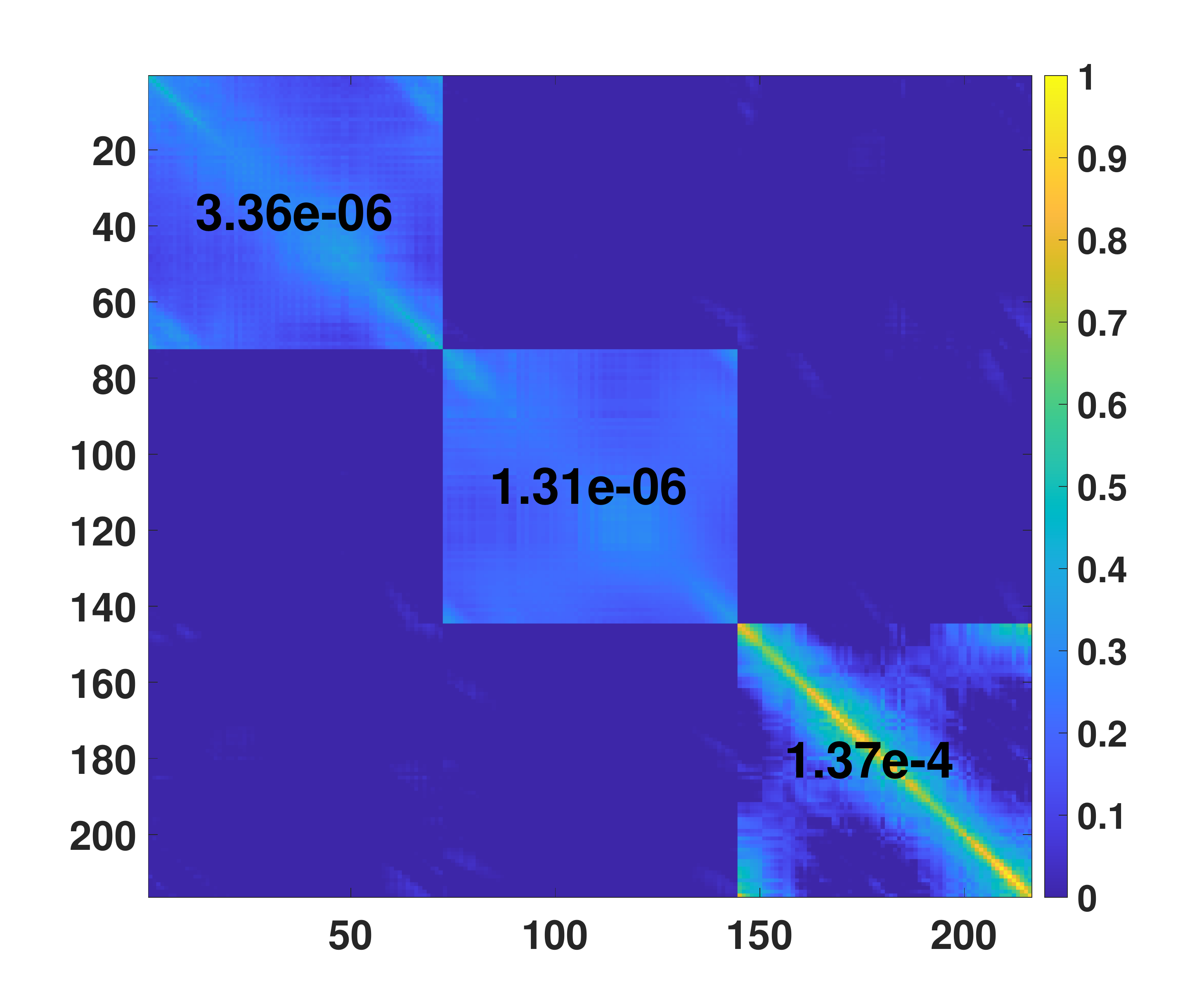}
		}
		\subfigure[$\|\mathbf{C}\|_*$]{
			\includegraphics [width=0.36\columnwidth]{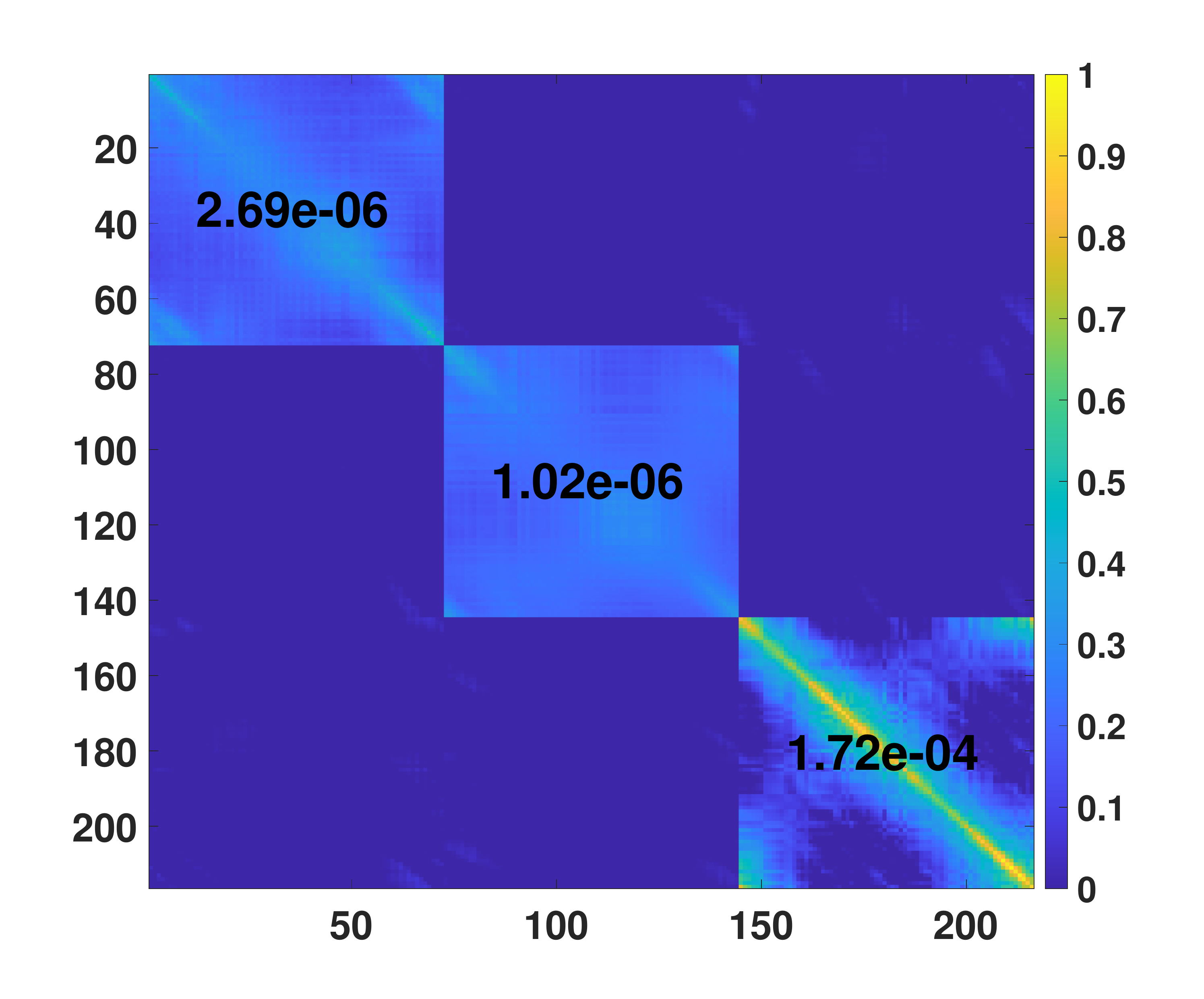}
		}
		\subfigure[Our]{
			\includegraphics [width=0.36\columnwidth]{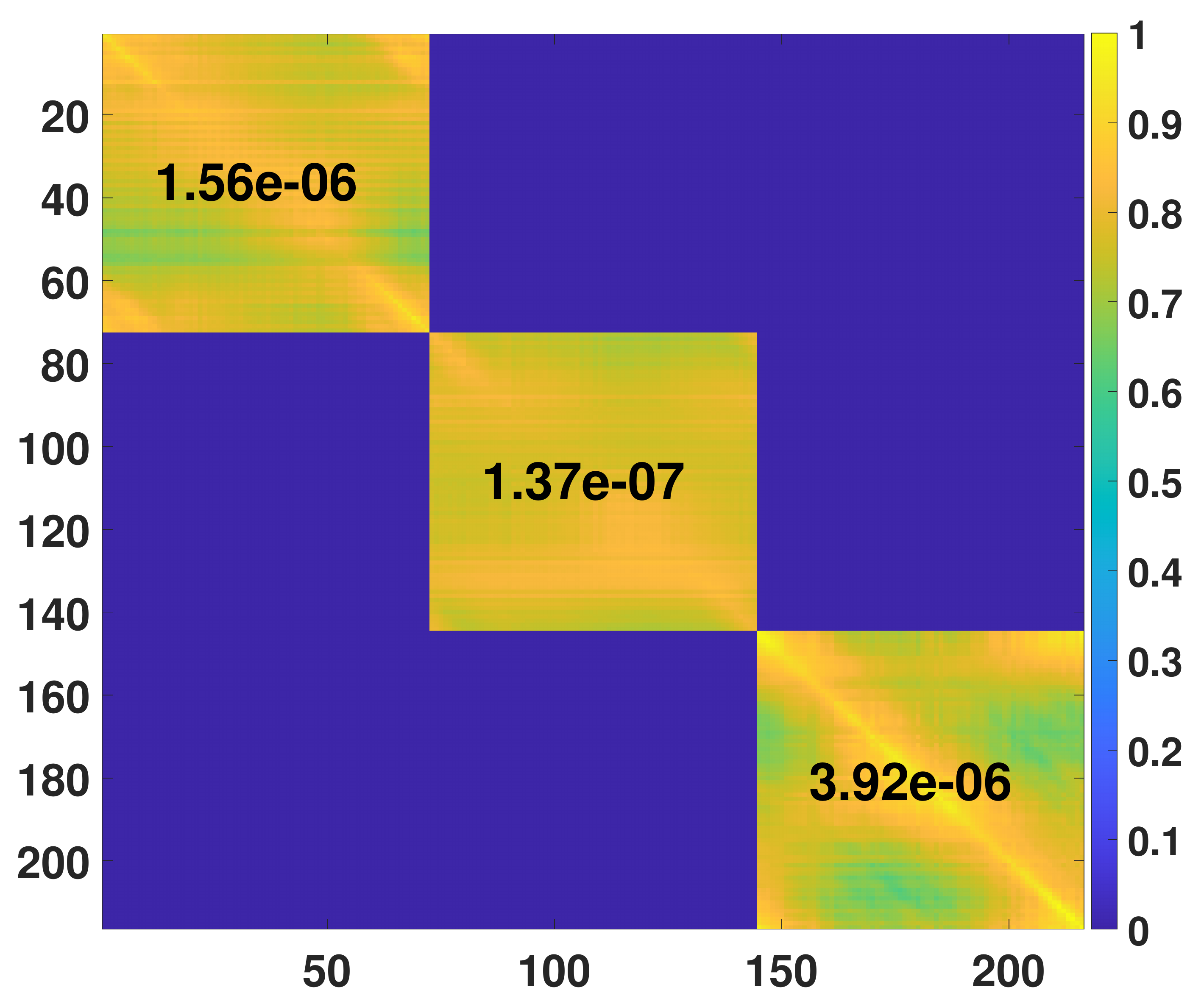}
		}	
		\caption{Qualitative comparison of different regularizers on the affinity matrix $\mathbf{C}$, where the number in the box is the $\textbf{variance}$ of the connection weight of each subspace.}
		\label{fig: different regularizer}
	\end{figure*}

	{To validate the significant performance of MESC-Net, we first generated one toy dataset, where three objects were chosen from COIL20 \cite{nene1996columbia}, and each object has 72 different views of size $32 \times 32$.} See the sample images of the toy dataset in Figure \ref{fig:Image}. We compared the learned affinity matrix of our method with those of DSC-Net \cite{ji2017deep} with different regularizers on $\mathbf{C}$, including $\|\mathbf{C}\|_1$, $\|\mathbf{C}\|_F^2$, and $\|\mathbf{C}\|_*$. {Figure \ref{fig: different regularizer} shows the learned affinity matrices with different regularizers, where the variance quantitatively displays the density-level of sub-matrices on $\mathbf{C}$, and a smaller value indicates a more uniform distribution of a block.} Note that we have normalized the data in each matrix by the min-max normalization.

\begin{figure*}
		\centering
		\subfigure[Ideal vs. Ideal]{
			\includegraphics [width=0.36\columnwidth]{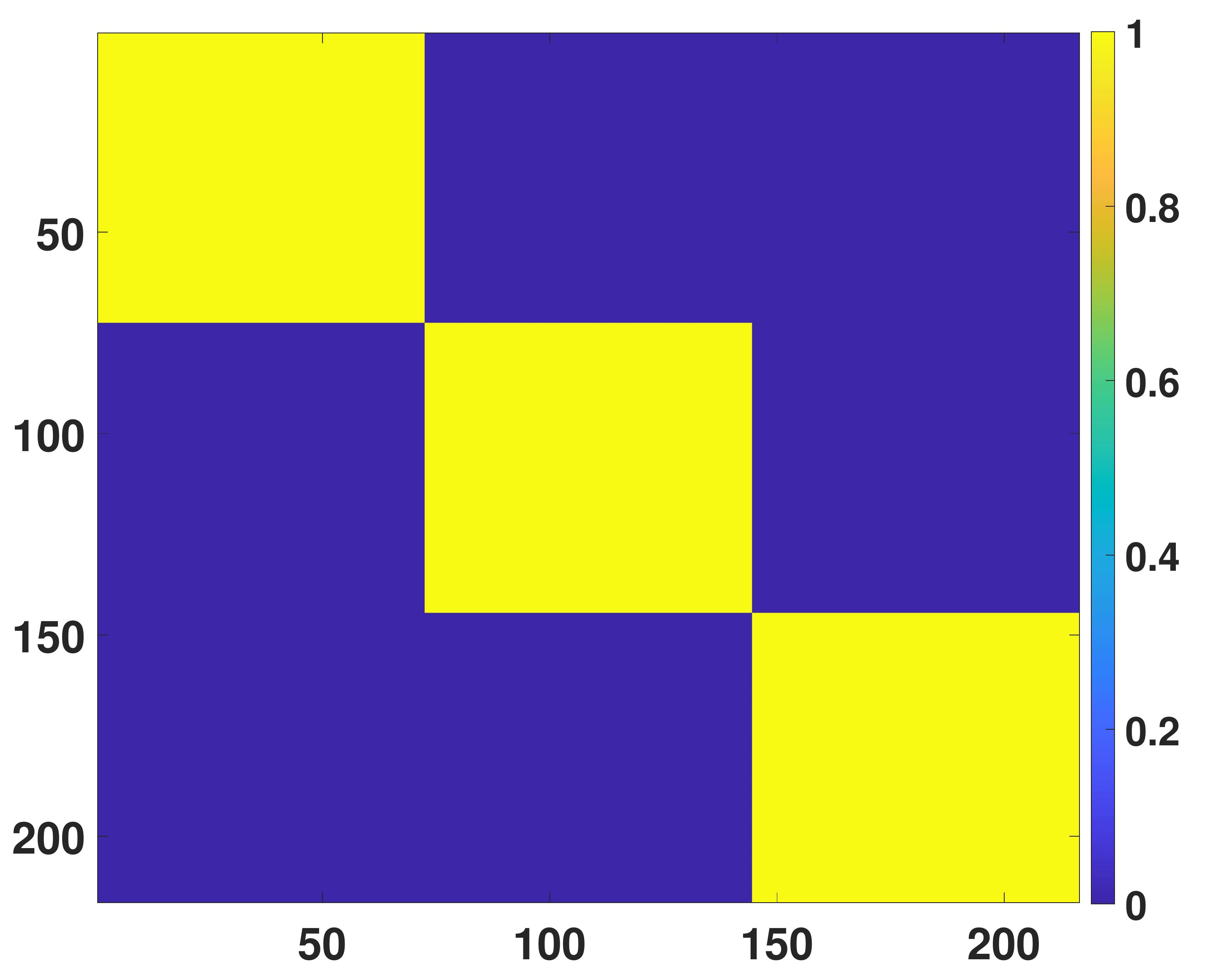}
		}
		\subfigure[Ideal vs. $\|\mathbf{C}\|_1$]{
			\includegraphics [width=0.36\columnwidth]{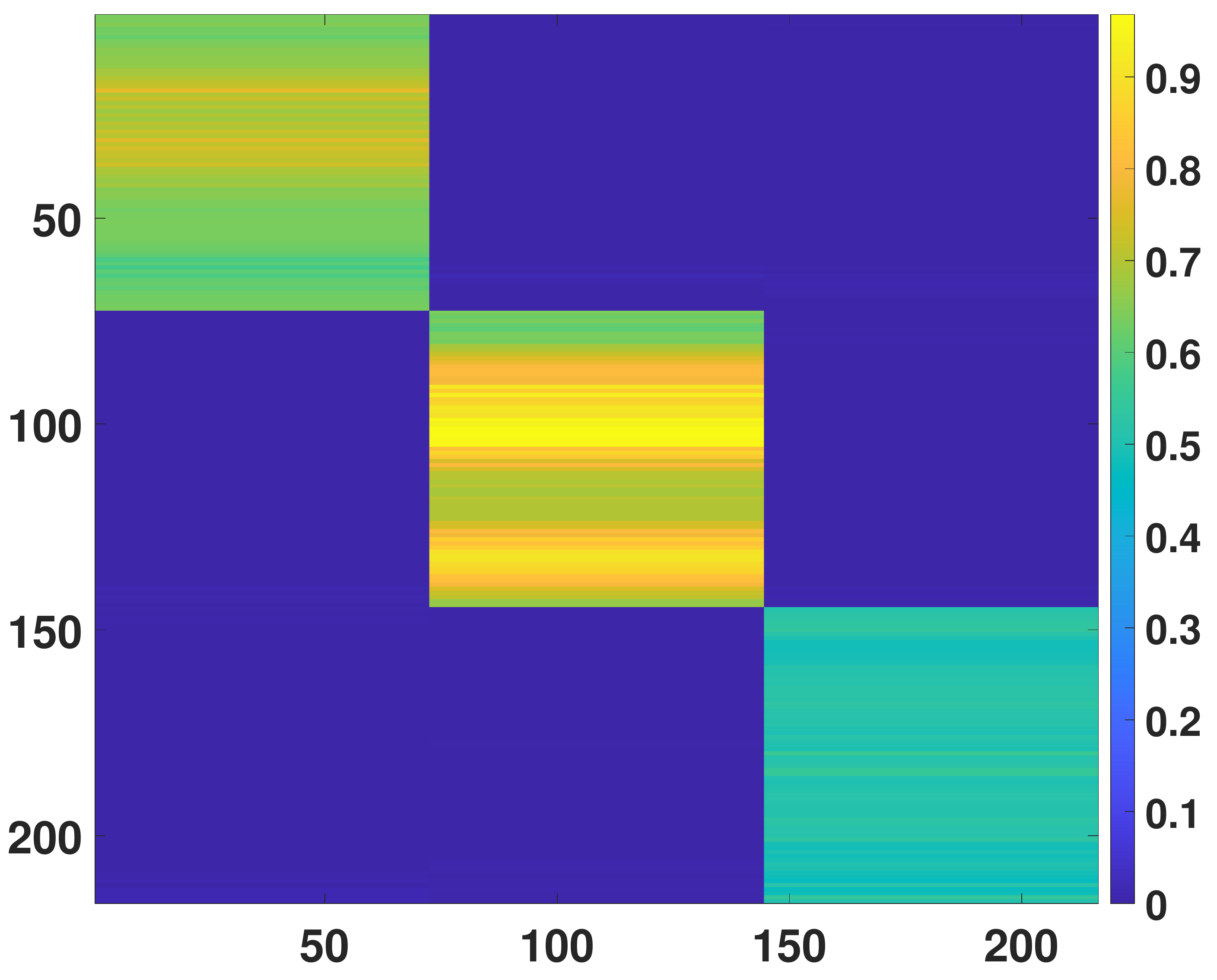}
		}
		\subfigure[Ideal vs. $\|\mathbf{C}\|_F^2$]{
			\includegraphics [width=0.36\columnwidth]{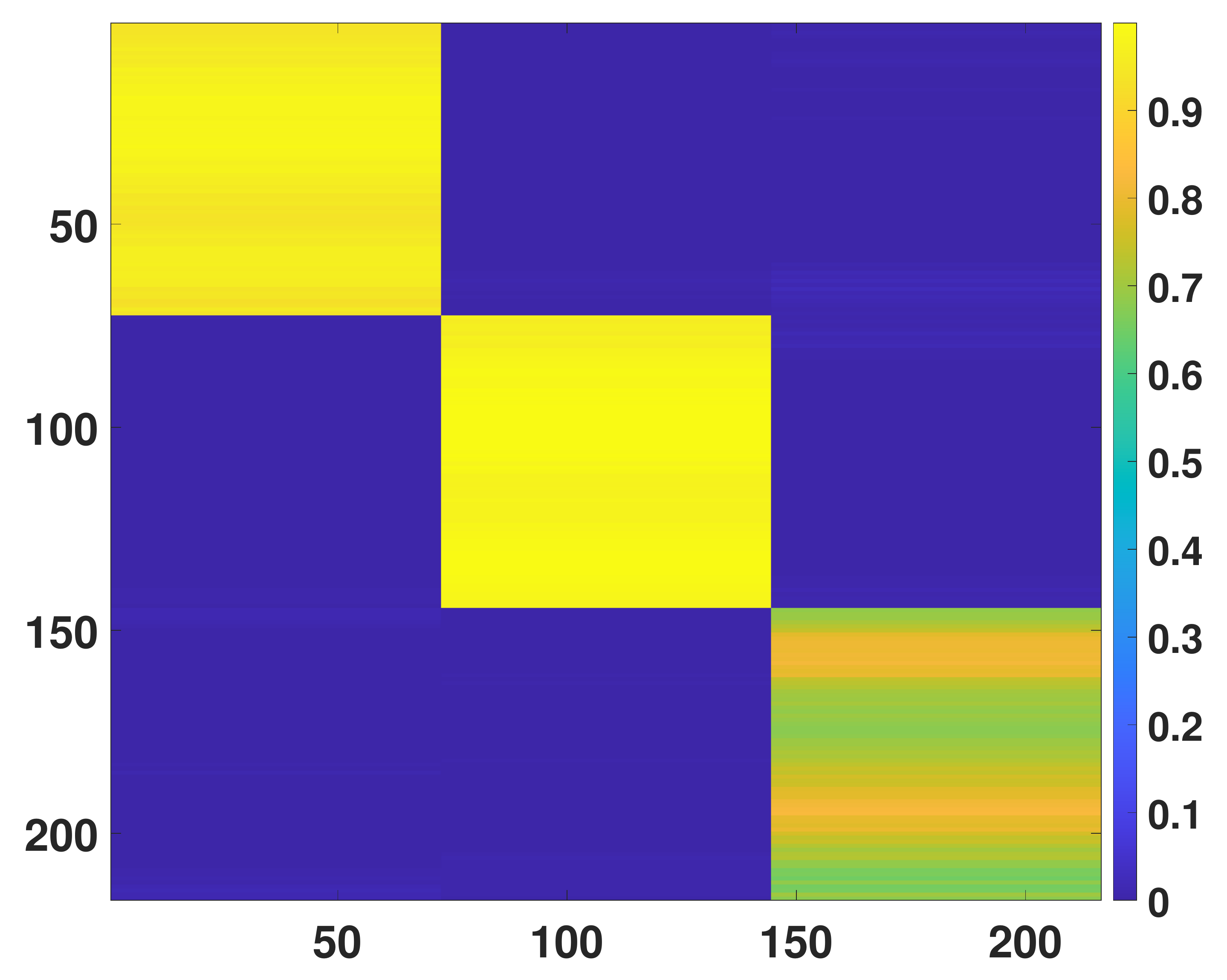}
		}
		\subfigure[Ideal vs. $\|\mathbf{C}\|_*$]{
			\includegraphics [width=0.36\columnwidth]{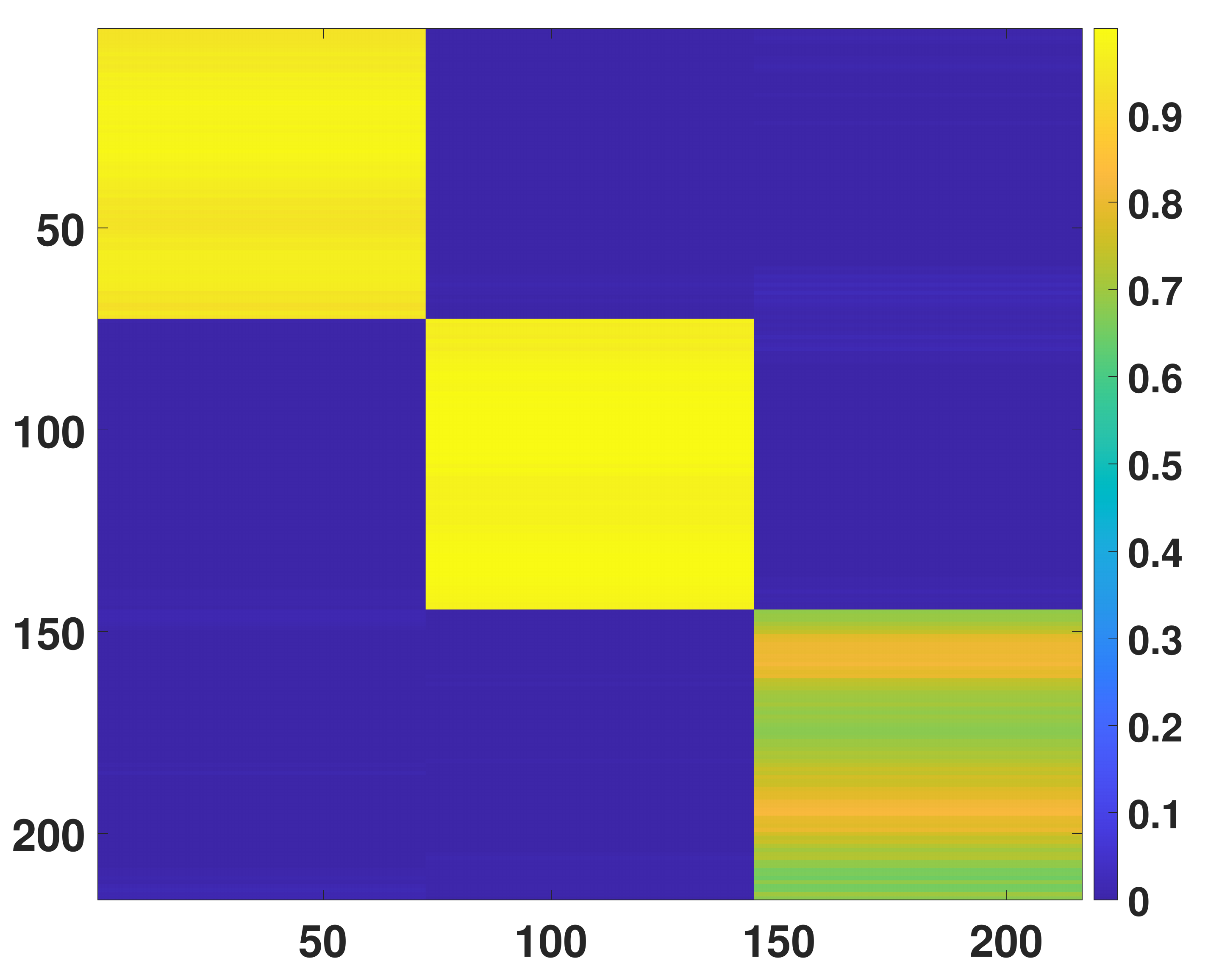}
		}
		\subfigure[Ideal vs. Our]{
			\includegraphics [width=0.36\columnwidth]{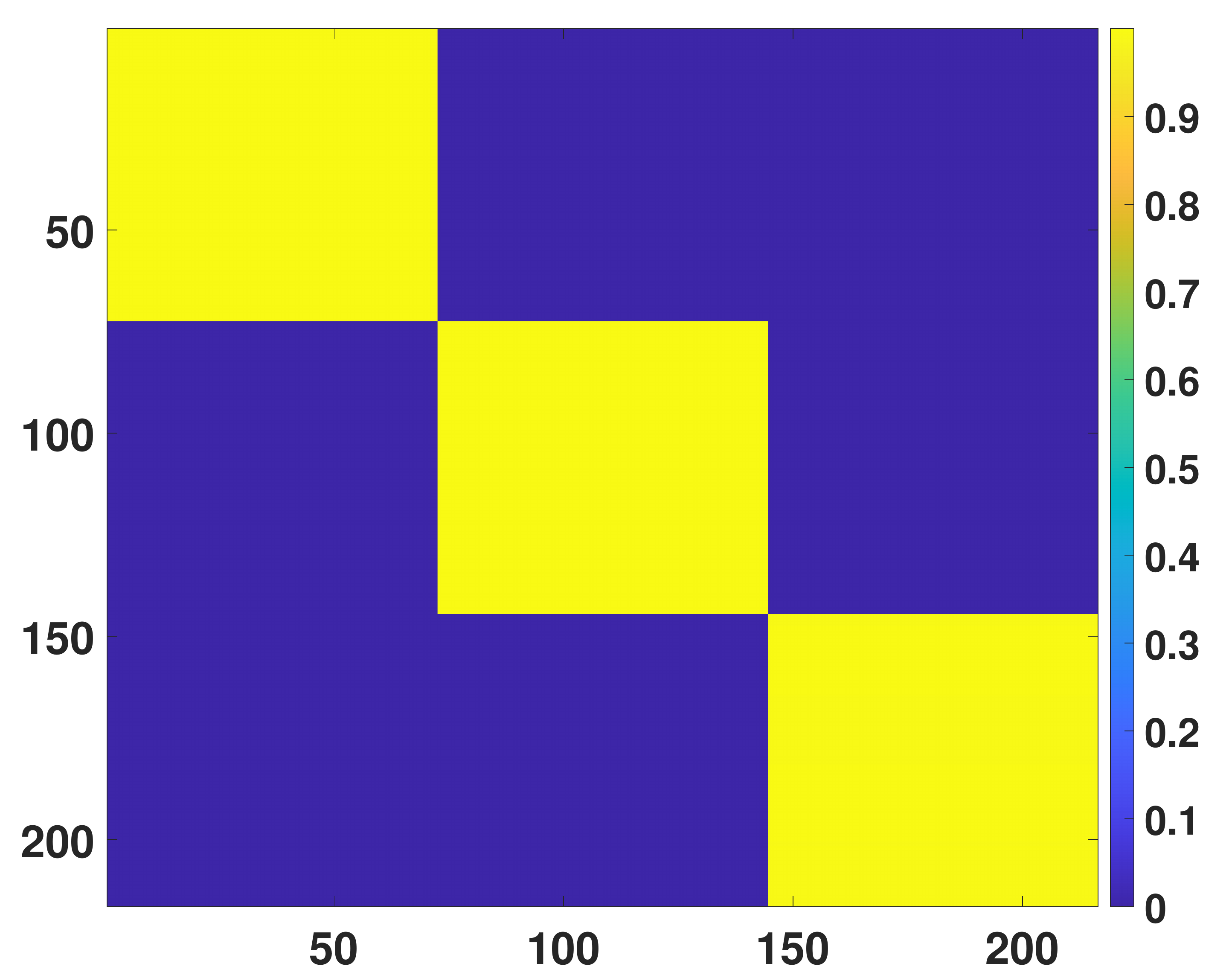}
		}	
		\caption{The cosine similarity between the ideal affinity matrix and others driven by different regularizers. Obviously, the affinity matrix driven by our method is the most similar to the ideal affinity matrix.}
		\label{fig: COSSIM}
	\end{figure*}

   From Figure \ref{fig: different regularizer}, we can observe that all the regularizers can generate a block diagonal matrix (i.e., satisfying the subspace-preserving property), but with significantly different appearances. For example, as shown in Figures \ref{fig: different regularizer} (b) - (d), the affinity matrix obtained by $\ell_1$ regularization is relatively sparse, the affinity matrix obtained by the nuclear norm is low-rank, and the affinity matrix obtained by the Frobenius norm becomes denser. \cite{dyer2013subspace,tang2016tri,liu2017learning} have validated the importance of the connectivity within each subspace. However, the Frobenius norm has a drawback that it will make the values in the subspace smaller (i.e., depressing the strength of connectivity), thereby reducing the discriminative ability to distinguish with the points in the off-diagonal blocks. In contrast, our model generates the densest connections in the subspace of the affinity matrix, and in each subspace, the values of the connections are strong and quite similar, as shown in Figure \ref{fig: different regularizer} (e). Furthermore, we computed the cosine similarity between the ideal matrix and the learned affinity matrices by different methods, which are shown in Figure \ref{fig: COSSIM}. Obviously, the affinity matrix driven by our method is quite close to the ideal affinity matrix. As will be shown in the next section, this will also lead to better clustering performance.
	

	\subsection{Block-diagonal Property Analysis}
	
	We first provide an important proposition about the block-diagonal property.
	
	\textbf{Proposition 1.} \cite{lu2018subspace} \emph{Consider a collection of data points drawn from $k$ independent subspaces $\left\lbrace \mathcal{S}_i \right\rbrace _{i=1}^k$ of dimensions $\left\lbrace d_i \right\rbrace _{i=1}^k$. Let $\mathbf{Z}_i\in\mathbb{R}^{d\times n_i}$ denote the data points in $ \mathcal{S}_i$, $rank(\mathbf{Z}_i)=d_i$, $\sum_{i=1}^{k}n_i=n$, and $\mathbf{Z}=[\mathbf{Z}_1,...,\mathbf{Z}_k]\in\Delta$, where $\Delta$ is a set consisting of matrices with nonzero columns. Considering the problem of Eq. (\ref{eq:SC}), assume that $\left\lbrace \mathbf{C}|\mathbf{Z}=\mathbf{Z}\mathbf{C} \cap\Omega\right\rbrace $ is nonempty and let $\mathbf{C}^*$ be any optimal solution, where $\Omega$ is some matrix set. \\
	If \textbf{conditions (a)} and \textbf{(b)} are satisfied, i.e., \\ 
		\textbf{(a)} $\quad f(\mathbf{C})=f(\mathbf{P}^\mathsf{T}\mathbf{C}\mathbf{P})$, for any permutation matrix $\mathbf{P}$, $\mathbf{P}^\mathsf{T}\mathbf{C}\mathbf{P}\in\Omega$ on $\left( \Omega, \Delta \right)$;\\
		\textbf{(b)} $\quad \mathbf{C}^*$ is the unique solution;\\
		we can conclude that $f(\mathbf{C})$ satisfies the block-diagonal property under the independent subspaces assumption.}
	
	{Then, we prove that our MESC-Net satisfies the above conditions of the block-diagonal property under the independent subspaces assumption.}

	\textbf{Proof of condition (a):} $H(\mathbf{C})$ is an entropy function, and Shannon \cite{shannon1948mathematical} showed that the entropy remains unchanged if the outcomes are re-ordered. This means that for any permutation matrix $\mathbf{P}$, $\mathbf{P}^\mathsf{T}\mathbf{C}\mathbf{P}\in\Omega$ on $\left( \Omega, \Delta \right)$, we have $H(\mathbf{P}^\mathsf{T}\mathbf{C}\mathbf{P}) = H(\mathbf{C})$, likewise that $f(\mathbf{C})$ (i.e., $-H(\mathbf{C})$) satisfies the condition (a) (i.e., $f(\mathbf{C})=f(\mathbf{P}^\mathsf{T}\mathbf{C}\mathbf{P})$, for any permutation matrix $\mathbf{P}$, $\mathbf{P}^\mathsf{T}\mathbf{C}\mathbf{P}\in\Omega$ on $\left( \Omega, \Delta \right)$). 
	
	\textbf{Proof of condition (b):} $f(\mathbf{C})$ can be rewritten as
	\begin{equation}
	\begin{aligned}
	& f(c_{i,j})\!=\!\lambda_1\sum_{i=1}^{n}\sum_{j=1}^{n} c_{i,j} \ln c_{i,j} \!+\!\lambda_2\!\left(\!z_{i,j}\!-\!\sum_{p=1}^{n} c_{i,p}z_{p,j}\!\right)\!\\ 
	& \rm{s.t.}, \quad c_{i,j} \geq 0.
	\label{eq:unique1}
	\end{aligned}
	\end{equation}	

	The first-order partial derivative of Eq. (\ref{eq:unique1}) with respect to (w.r.t.) $c_{i,j}$ is
	\begin{equation}
	\begin{aligned}
	& \frac{\partial f(c_{i,j})}{\partial c_{i,j}} = \lambda_1\left( \ln c_{i,j} + 1 \right) - \lambda_2 z_{j,j}\\ 
	& \rm{s.t.}, \quad c_{i,j} > 0,
	\label{eq:unique2}
	\end{aligned}
	\end{equation}
    and the corresponding second-order partial derivative is,
	\begin{equation}
	\begin{aligned}
	& \nabla^2 f = \frac{\partial^2 f(c_{i,j})}{\partial c_{i,j}^2} = \frac{\lambda_1}{c_{i,j}} > 0 \\
	& \rm{s.t.}, \quad c_{i,j} > 0.
	\label{eq:unique3}
	\end{aligned}
	\end{equation}
	
	According to Eq. (\ref{eq:unique3}), $f$ is strictly convex as $\nabla^2 f>0$ for all $c_{i,j}\in \mathbf{dom} f$ \cite{boyd2004convex}. As a strictly convex function only has one global minimum \cite{soriano1993global}, the solution of $f(\mathbf{C})$ (i.e., $\mathbf{C^*}$) is the unique solution.

	Therefore, we can conclude that our method satisfies the block-diagonal property. That is, when input data points are drawn from $k$ independent subspaces, we have
	\begin{equation}
	\mathbf{C}^*=
	\begin {bmatrix}
	\mathbf{C}_1^*	& 0 			&\cdots			&0 \\
	0				& \mathbf{C}_2^*&\cdots			&0 \\
	\vdots			& \vdots		&\ddots			&\vdots \\
	0				& 0				&\cdots			&\mathbf{C}_k^*
	\end {bmatrix},
	\label{eq:Z*}
	\end{equation}
	with $\mathbf{C}_i^*\in\mathbb{R}^{n_i\times n_i}$ corresponding to the sub-matrix $\mathbf{Z}_i$.
	
	\begin{table*}
	\caption{Description of the adopted datasets (i.e., Toy, ORL, COIL20, COIL40, COIL100, EYaleB, USPS and MNIST) and the corresponding network structures.}
	\label{tab:datasets and NN setting}
		\centering
		\begin{tabular}{l|ccc|c|cccccc}
			\hline\hline
			Dataset				& \# samples 			&\# features 					&\# subjects 		
			& Layers 		& \multicolumn{3}{c}{\# encoder} 		& \multicolumn{3}{c}{\# decoder} \\
			\hline
			\multirow{2}{*}{Toy} & \multirow{2}{*}{216} &\multirow{2}{*}{ 32$\times$32} & \multirow{2}{*}{3} 
			& kernel size 	& \multicolumn{3}{c}{3$\times$3}  & \multicolumn{3}{c}{3$\times$3}	\\
			& & & 
			& channels 		& \multicolumn{3}{c}{15}  		& \multicolumn{3}{c}{15}\\
			\hline
			\multirow{2}{*}{ORL \cite{samaria1994parameterisation}} & \multirow{2}{*}{400} &\multirow{2}{*}{ 32$\times$32} & \multirow{2}{*}{40} 
			& kernel size 	& 3$\times$3 & 3$\times$3  &3$\times$3  & 3$\times$3  & 3$\times$3  & 3$\times$3 \\
			& & & 
			& channels 		& 3 		 & 3 		   & 5   		& 5  		  & 3  			& 3 \\
			\hline	
			\multirow{2}{*}{COIL20 \cite{zhou2018deep}} & \multirow{2}{*}{1440} &\multirow{2}{*}{ 32$\times$32} & \multirow{2}{*}{20} 
			& kernel size 	& \multicolumn{3}{c}{3$\times$3}  & \multicolumn{3}{c}{3$\times$3} \\
			& & & 
			& channels 		& \multicolumn{3}{c}{15}  		& \multicolumn{3}{c}{15}  \\
			\hline	
			\multirow{2}{*}{COIL40 \cite{zhou2018deep}} & \multirow{2}{*}{2880} &\multirow{2}{*}{ 32$\times$32} & \multirow{2}{*}{40} 
			& kernel size 	& \multicolumn{3}{c}{3$\times$3}  & \multicolumn{3}{c}{3$\times$3}	\\
			& & & 
			& channels 		& \multicolumn{3}{c}{20}  		& \multicolumn{3}{c}{20} \\
			\hline	
			\multirow{2}{*}{COIL100 \cite{nene1996columbia}} & \multirow{2}{*}{7200} &\multirow{2}{*}{ 32$\times$32} & \multirow{2}{*}{72} 
			& kernel size 	& \multicolumn{3}{c}{5$\times$5}  & \multicolumn{3}{c}{5$\times$5}	\\
			& & & 
			& channels 		& \multicolumn{3}{c}{50}  		& \multicolumn{3}{c}{50} \\
			\hline	
			\multirow{2}{*}{EYaleB \cite{georghiades2001few}} & \multirow{2}{*}{2432} &\multirow{2}{*}{ 48$\times$42} & \multirow{2}{*}{38}
			& kernel size 	& 5$\times$5 & 3$\times$3  &3$\times$3  & 3$\times$3  & 3$\times$3  & 5$\times$5\\
			& & & 
			& channels 		& 10 		 & 20 		   & 30   		& 30  		  & 20  		& 10 \\
			\hline
			\multirow{2}{*}{USPS \cite{hull1994database}} & \multirow{2}{*}{9298} &\multirow{2}{*}{ 16$\times$16} & \multirow{2}{*}{10} 
			& kernel size 	& 5$\times$5 & 3$\times$3  &3$\times$3  & 3$\times$3  & 3$\times$3  & 5$\times$5  \\
			& & & 
			& channels 		& 10 		 & 20 		   & 30   		& 30  		  & 20  		& 10  \\
			\hline
			\multirow{2}{*}{MNIST \cite{lecun1998gradient}} & \multirow{2}{*}{10000} &\multirow{2}{*}{ 28$\times$28} & \multirow{2}{*}{10} 
			& kernel size 	& 5$\times$5 & 3$\times$3  &3$\times$3  & 3$\times$3  & 3$\times$3  & 5$ \times $5 \\
			& & & 
			& channels 		& 10 		 & 20 		   & 30   		& 30  		  & 20  		& 10\\
			\hline\hline
		\end{tabular}
	\end{table*}

	\section{Experiments}

	\subsection{Datasets}
	We conducted the experiments on seven commonly used benchmark datasets, including: ORL\footnote{http://www.cl.cam.ac.uk/research/dtg/attarchive/facedatabase.html}, COIL20\footnote{http://www.cs.columbia.edu/CAVE/software/softlib/coil-20.php}, COIL40\footnote{https://github.com/sckangz/L2SP}, COIL100\footnote{https://www1.cs.columbia.edu/CAVE/software/softlib/coil-100.php}, EYaleB\footnote{http://vision.ucsd.edu/ iskwak/ExtYaleDatabase/ExtYaleB.html}, USPS\footnote{https://ribbs.usps.gov/index.cfm?page=address\_info\_systems} and MNIST\footnote{http://yann.lecun.com/exdb/mnist}.
	
	\begin{itemize}
		\item	\textbf{ORL}. The Olivetti Research Ltd database contains 400 images of 40 individuals, in which these images were captured at different times and with different lighting conditions, and each image has $32 \times 32$ pixels. As \cite{ji2017deep} said, this dataset is challenging for subspace clustering because (i) the face subspaces have more non-linearity due to varying facial expressions and details; (ii) the dataset size is much smaller.
		\item	\textbf{COIL20, COIL40, COIL100}. The Columbia Object Image Library databases contains three variants: COIL20 (i.e., 1440 images of 20 objects), COIL40 (i.e., 2880 images of 40 objects) and COIL100 (i.e., 7200 images of 100 objects), with 72 different views. And each image has $32 \times 32$ pixels.
		\item	\textbf{EYaleB}. The Extended Yale B dataset consists of 38 subjects and each subject has 64 face images with size $192 \times 168$ acquired under various pose and lighting conditions. Following the setting of \cite{elhamifar2013sparse}, each image was down-sampled to $48 \times 42$ pixels.
		\item 	\textbf{USPS}. The United States Postal Service database includes ten classes (i.e., `0'–`9') of 11000 handwritten digits. We used a popular subset containing 9298 handwritten digit images for the experiments, and all of these images were normalized to $16 \times 16$.
		\item	\textbf{MNIST}. The Modified National Institute of Standards and Technology database has a training set of 60000 examples and a test set of 10000 examples. We used the 10000 test samples to conduct the experiment, and each image has $28 \times 28$ pixels.
	\end{itemize}

	{The brief of the used datasets and the corresponding network structures are summarized in Table \ref{tab:datasets and NN setting}.}
	
   	\begin{figure}
		\centering
		\subfigure[]{
			\includegraphics [width=0.46\columnwidth]{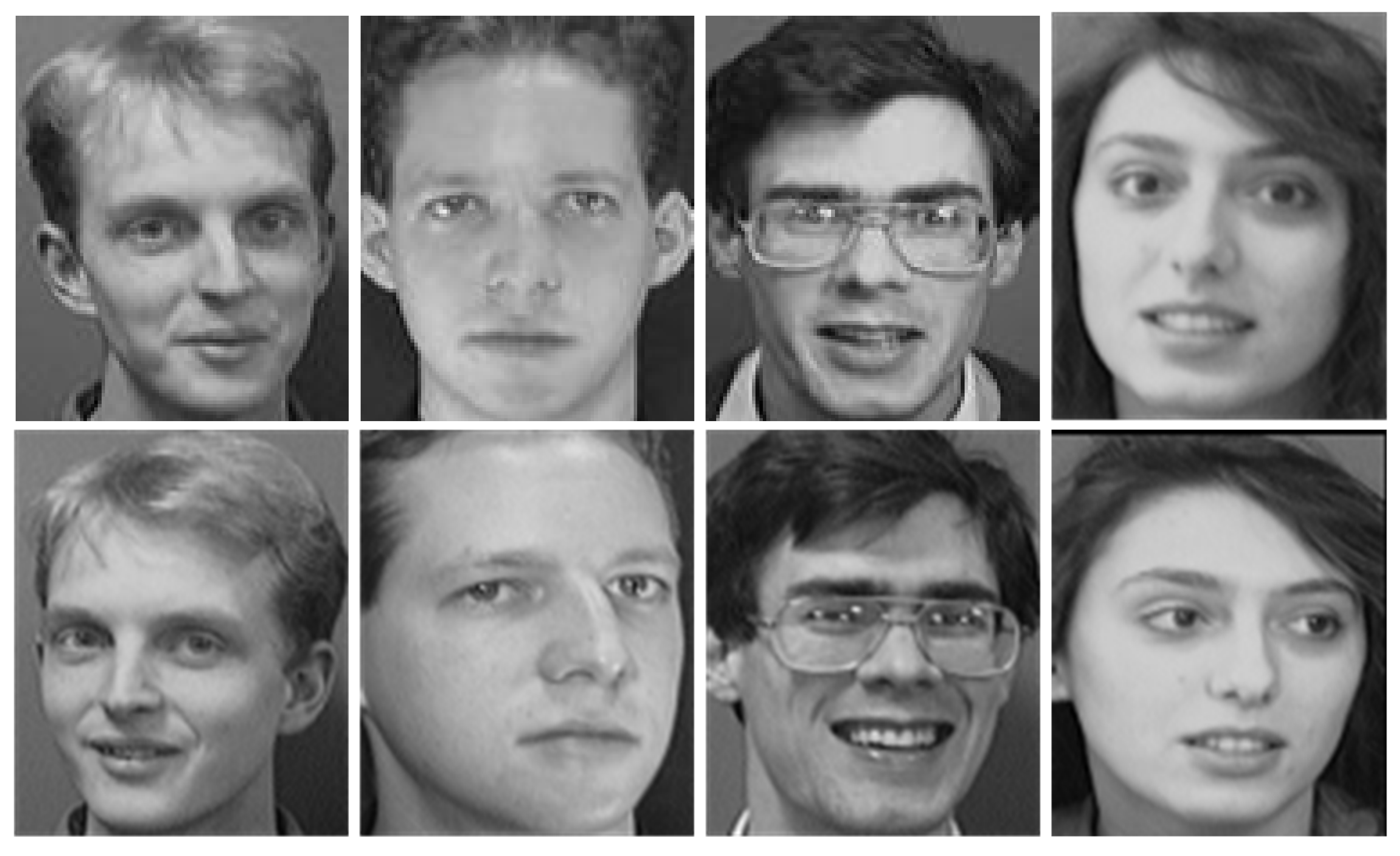}
		}
		\subfigure[]{
			\includegraphics [width=0.46\columnwidth]{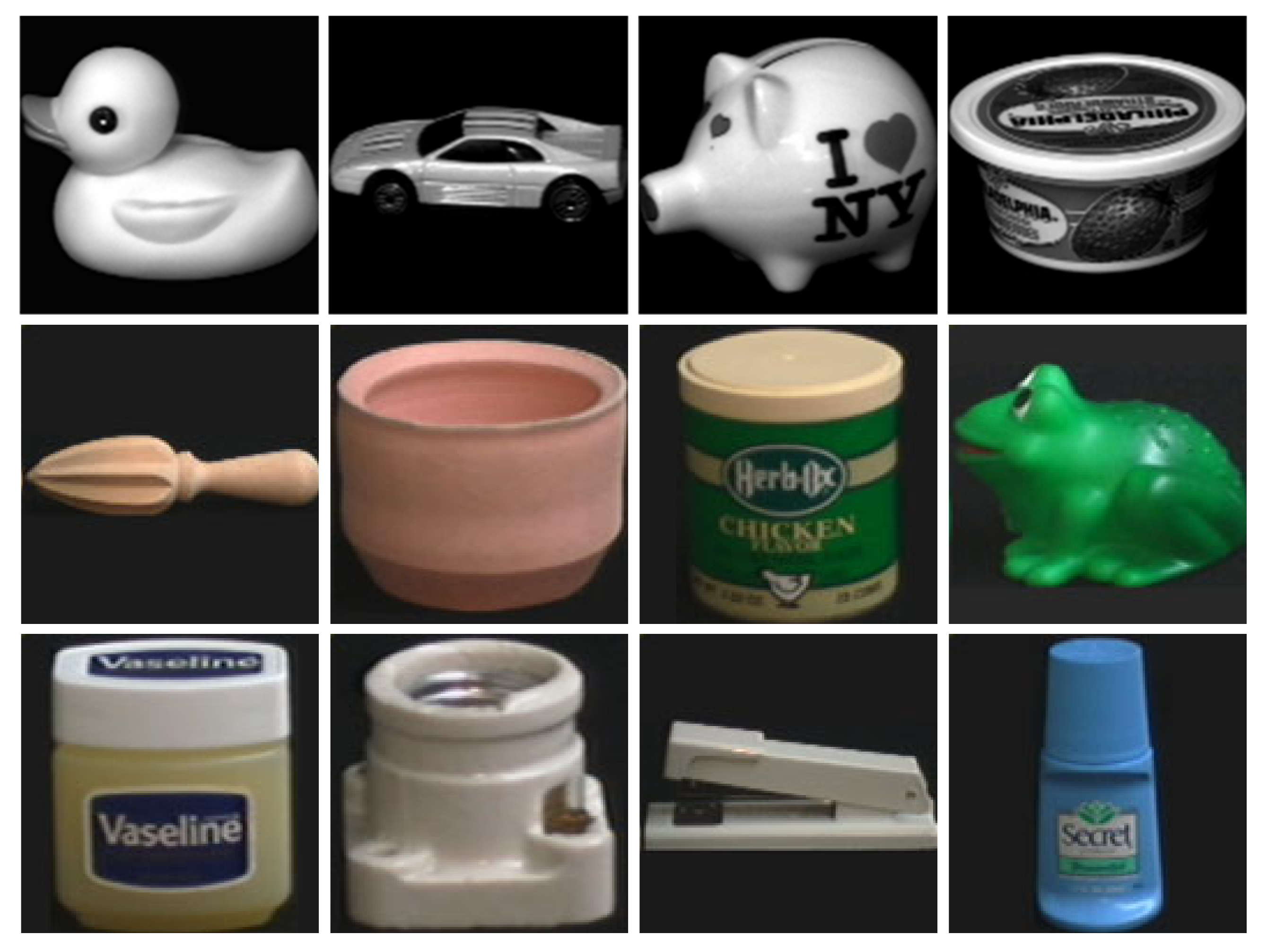}
		}
		\subfigure[]{
			\includegraphics [width=0.46\columnwidth]{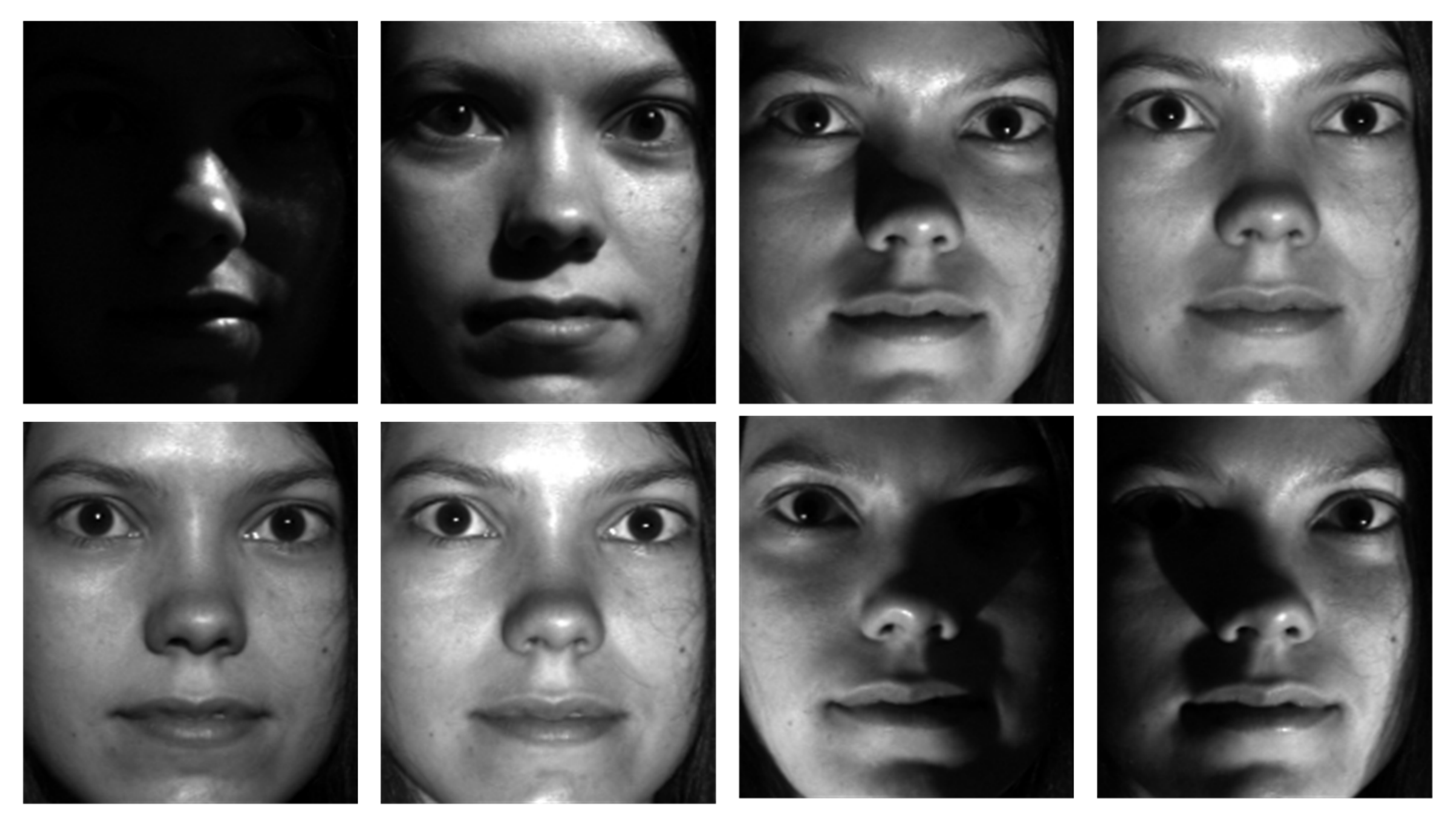}
		}
		\subfigure[]{
			\includegraphics [width=0.46\columnwidth]{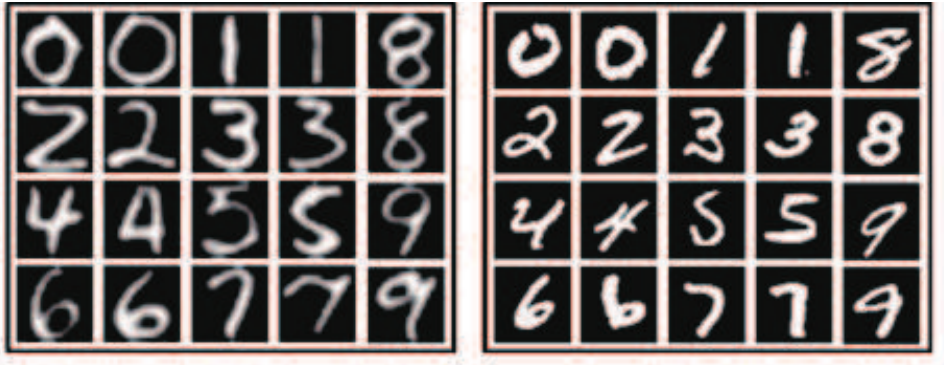}
		}
		\caption{{Sample images from seven datasets used in the experiments. (a) ORL. (b) COIL20, COIL40 and COIL100. (c) EYaleB. (d) USPS and MNIST.}}
		\label{fig: Matrix}
	\end{figure}
	
	\subsection{Compared Methods} 

	We compared our method with five conventional subspace clustering methods and seven deep subspace clustering methods:

	\begin{table*}
    \caption{Comparison of ACC (\%) of different methods on seven datasets. The best and second best results are highlighted with \textbf{bold} and \underline{underline}, respectively, and `-' indicates that the corresponding results are unavailable in the original papers.}
    \centering
	\label{tab: ACC}
        \begin{tabular}{c|c|c|c|c|c|c|c}
        \hline
        Datasets                                           & ORL               & COIL20            & COIL40            & COIL100           & EYaleB            & USPS              & MNIST             \\ \hline
        SSC \cite{elhamifar2009sparse,elhamifar2013sparse} & 70.50             & 85.17             & 71.91             & 55.10             & 70.64             & -                 & -                 \\
        KSSC \cite{patel2014kernel}                        & 65.75             & 75.35             & 65.49             & 52.82             & 72.25             & -                 & -                 \\
        SSC-OMP \cite{you2016scalable}                     & 62.95             & 70.14             & 44.31             & 32.71             & 76.48             & -                 & -                 \\
        LRR \cite{liu2010robust,liu2012robust}             & 66.50             & 69.79             & 64.93             & 46.82             & 64.88             & -                 & -                 \\
        LRSC \cite{vidal2014low}                           & 67.50             & 68.75             & 63.27             & 49.33             & 70.11             & -                 & -                 \\ \hline
        LRAE \cite{chen2018subspace}                       & 82.58             & -                 & -                 & 56.62             & -                 & -                 & 62.07             \\
       DASC \cite{zhou2018deep}                          & 88.25             & 96.39             & 83.54             & -                 & \textbf{98.56}    & -                 & -                 \\
        DKM \cite{fard2020deep}                            & 46.82             & 66.51             & 17.13             & 51.96             & 17.13             & 75.70             & {72.72}           \\
        DSLSP-L1 \cite{kang2020structure}                  & 87.00             & 97.43             & 83.89             & 65.86             & {97.57}           & \underline{83.29} & {78.12}           \\
        DSLSP-L2 \cite{kang2020structure}                  & {87.75}           & {97.57}           & {84.17}           & 65.54             & {97.62}           & {83.18}           & {78.70}           \\
        DSC-Net-L1 \cite{ji2017deep}                       & 85.75             & 93.05             & 80.03             & 60.67             & 96.67             & 79.65             & 73.44             \\
        DSC-Net-L2 \cite{ji2017deep}                       & 86.00             & 94.86             & 80.75             & {67.71}           & 97.33             & 77.64             & {73.61}           \\ \hline
        Our w/o Net                                        & \underline{89.75} & \textbf{98.19}    & \underline{89.13} & \underline{69.97} & \underline{98.03} & \textbf{83.53}    & \underline{80.28} \\
        Our                                                & \textbf{90.25}    & \underline{98.13} & \textbf{89.51}    & \textbf{71.88}    & {97.45}           & {81.49}           & \textbf{81.11}    \\ \hline
        \end{tabular}
	\end{table*}
	
		\begin{table*}
		\caption{Comparison of NMI (\%) of different methods on seven datasets. the best and second-best results are highlighted with \textbf{bold} and \underline{underline}, respectively, and `-' indicates that the corresponding results are unavailable in the original papers.}
		\centering
		\label{tab: NMI}
            \begin{tabular}{c|c|c|c|c|c|c|c}
            \hline
            Datasets                                           & ORL            & COIL20            & COIL40            & COIL100           & EYaleB            & USPS              & MNIST             \\ \hline
            SSC \cite{elhamifar2009sparse,elhamifar2013sparse} & 84.59          & 88.92             & 82.12             & 58.41             & 77.96             & -                 & -                 \\
            KSSC \cite{patel2014kernel}                        & 80.70          & 82.43             & 78.88             & 60.47             & 73.59             & -                 & -                 \\
            SSC-OMP \cite{you2016scalable}                     & 79.52          & 74.12             & 65.45             & 67.56             & 78.03             & -                 & -                 \\
            LRR \cite{liu2010robust,liu2012robust}             & 86.03          & 87.47             & 78.28             & 47.21             & 86.36             & -                 & -                 \\
            LRSC \cite{vidal2014low}                           & 81.56          & 84.52             & 77.37             & 58.10             & 82.64             & -                 & -                 \\ \hline
            LRAE \cite{chen2018subspace}                       & 91.26          & -                 & -                 & 79.87             & -                 & -                 & 65.49             \\
            DASC \cite{zhou2018deep}                           & 93.15          & 96.86             & 91.96             & -                & \textbf{98.01}    & -                 & -                 \\
            DKM \cite{fard2020deep}                            & 73.32          & 79.71             & 78.40             & 77.72             & 27.04             & 77.60             & {66.39}           \\
            DSLSP-L1 \cite{kang2020structure}                  & 92.37          & 97.31             & 92.62             & 89.14             & 96.68             & {83.70}           & {80.43}           \\
            DSLSP-L2 \cite{kang2020structure}                  & {92.49}        & {97.40}           & {92.67}           & {89.39}           & {96.74}           & 83.49             & \underline{81.15} \\
            DSC-Net-L1 \cite{ji2017deep}                       & 90.23          & 93.53             & 88.52             & 85.86             & 96.87             & 82.95             & 74.94             \\
            DSC-Net-L2 \cite{ji2017deep}                       & 90.34          & 94.08             & 89.41             & 89.08             & {97.03}           & 78.86             & 75.15             \\ \hline
            Our w/o Net                                   & \underline{93.23}   & \textbf{98.19}    & \textbf{95.40}    & \underline{89.86} & \underline{97.27} & \underline{85.80} & {80.27}           \\
            Our                                                & \textbf{93.59} & \underline{98.17} & \underline{95.39} & \textbf{90.76}    & {96.61}           & \textbf{86.34}    & \textbf{82.26}    \\ \hline
            \end{tabular}
	\end{table*}

	\begin{itemize}
		\item	\textbf{SSC} \cite{elhamifar2013sparse} produces subspace-preserving data affinity by expressing each data point as a sparse linear combination of the other data points from the same subspace.
		\item	\textbf{KSSC} \cite{patel2014kernel} extends SSC to non-linear manifold by the kernel trick.
		\item	\textbf{SSC-OMP} \cite{you2016scalable} is a scalable sparse subspace clustering method by introducing the orthogonal matching pursuit.
		\item	\textbf{LRR} \cite{liu2012robust} seeks the lowest rank representation among all the candidates that can represent the data samples as linear combinations of the bases in a given dictionary.
		\item	\textbf{LRSC} \cite{vidal2014low} introduces a general optimization framework for solving the subspace clustering problem in the case of data corrupted by noise or gross errors.
		
		\item   \textbf{LRAE} \cite{chen2018subspace} combines the advantages of LRR and the auto-encoder.
		\item   \textbf{DASC} \cite{zhou2018deep} combines the advantages of adversarial learning and the auto-encoder.
		\item	\textbf{DKM} \cite{fard2020deep} learns a deep representation and performs K-means jointly to achieve better clustering performance. 
		\item	\textbf{DSLSP-L1}, \textbf{DSLSP-L2} \cite{kang2020structure} propose a deep structure learning framework by the $\ell_1$ norm and the $\ell_2$ norm that retains the pairwise similarities between the data points.
		\item   \textbf{DSC-Net-L1}, \textbf{DSC-Net-L2} \cite{ji2017deep} design a fully connected layer between the encoder and the decoder to simulate the self-expressiveness property and consider two kinds of regularizations (i.e., the $\ell_1$ norm and the $\ell_2$ norm) on affinity matrix.
		\item   \textbf{Our w/o Net} applies the proposed ME regularization on the network architecture of DSC-Net \cite{ji2017deep}.
		\item   \textbf{Our} denotes the proposed method, which applies the ME regularization on the designed decoupling framework.
	\end{itemize}

	For these approaches, we used the codes released by authors and obtained the best results on each dataset by carefully tuning their parameters. Since the codes of LRAE and DASC are not publicly available, we only reported the performances provided in the original paper \cite{chen2018subspace,zhou2018deep}. Due to the memory and computation issue, the experiments of the conventional subspace clustering methods on USPS and MNIST datasets were not conducted. 
	
	{Notably, the only difference among `DSC-Net-L1', `DSC-Net-L2', and `Our w/o Net' is the adopted regularization. The only difference between `Our w/o Net' and `Our' is the adopted deep network framework. From these settings, it is easy to validate the effectiveness and efficiency of the proposed ME regularization and the decoupling architecture.}

	\subsection{Evaluation Metrics}
	We used two popular clustering metrics (i.e., accuracy (ACC)) and normalized mutual information (NMI), to evaluate the clustering performance of all the methods. Specifically, ACC is defined as 
	\begin{equation}
	\begin{aligned}
	& \text{ACC} \left( \%\right)  =  \frac{\sum_{i=1}^{n} \mathbf{1} \left\lbrace y_i = m(c_i) \right\rbrace }{n} \times 100\%,
	\label{eq:ACC}
	\end{aligned}
	\end{equation}
	where $y_i$ is the ground-truth label, $c_i$ is the clustering assignment, $m(\cdot)$ enumerates the mapping between clustering assignments and ground-truth labels by the Kuhn-Munkres algorithm \cite{munkres1957algorithms}, and $\mathbf{1}\left\lbrace \cdot \right\rbrace$ is the indicator function returning 1 or 0. NMI is defined as
	\begin{equation}
	\begin{aligned}
	& \text{NMI}\left( \%\right)=\frac{\sum_{y\in Y, \hat{y} \in \hat{Y}} p(y,\hat{y})\!\log\!\left( \!\frac{p(y,\hat{y})}{p(y)p(\hat{y})}\!\right)}{\max\left( H(Y), H(\hat{Y}) \right) } \!\times \!100\%,
	\label{eq:NMI}
	\end{aligned}
	\end{equation}
	where $Y$ and $\hat{Y}$ denote two sets of clusters, $y$ and $\hat{y}$ are the corresponding labels, $p(y)$ and $p(\hat{y})$ are the marginal probability mass functions of $Y$ and $\hat{Y}$ respectively, $p(y,\hat{y})$ is the joint probability mass function of $Y$ and $\hat{Y}$, and $H(\cdot)$ represents the entropy function \cite{shannon1948mathematical}. Note that the larger the values of ACC and NMI, the better the clustering performance.

	\begin{figure*}[ht]
		\centering
		\subfigure[DSC-Net-L1]{
			\includegraphics [width=0.58\columnwidth]{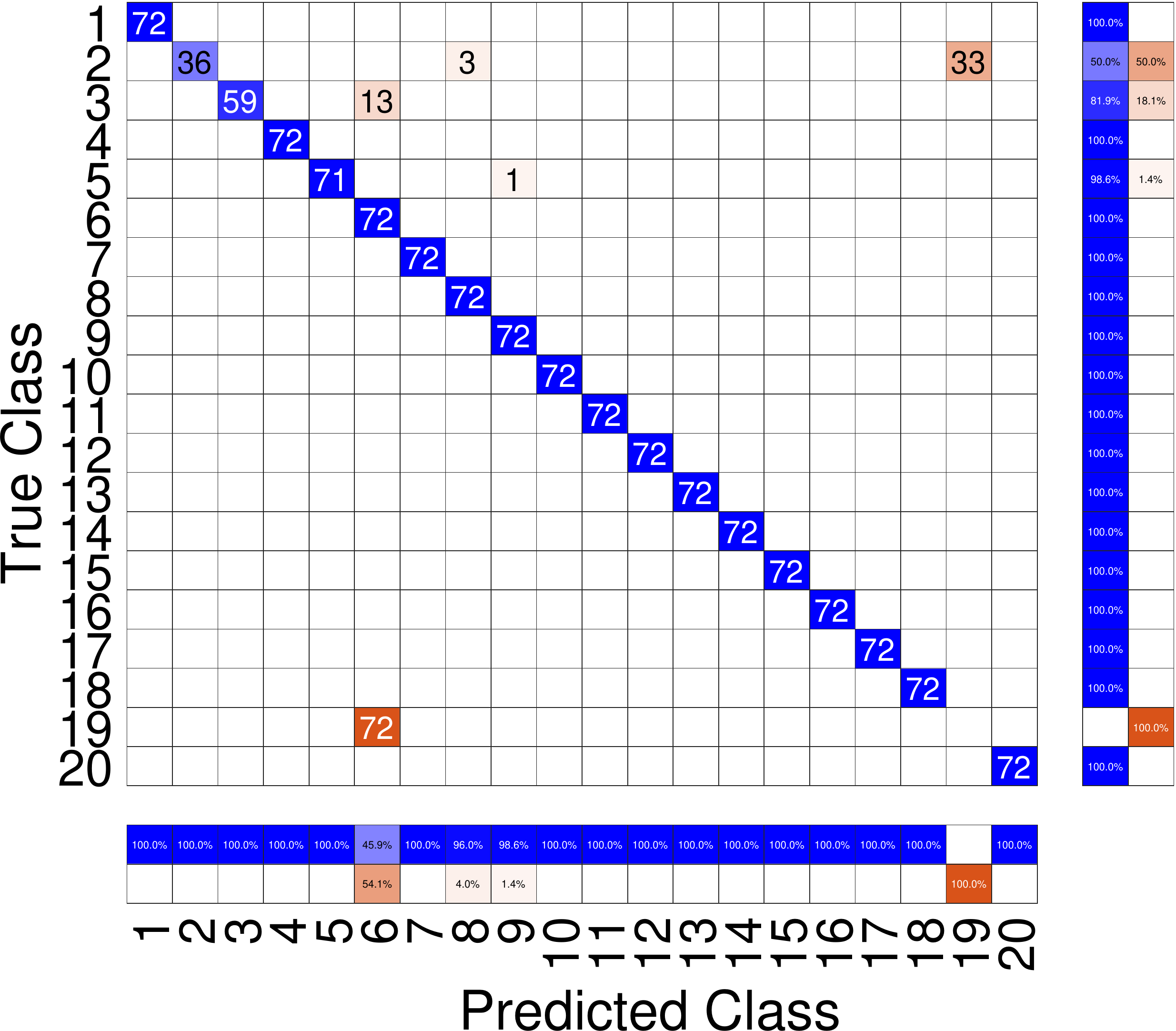}
		}
		\subfigure[DSC-Net-L2]{
			\includegraphics [width=0.58\columnwidth]{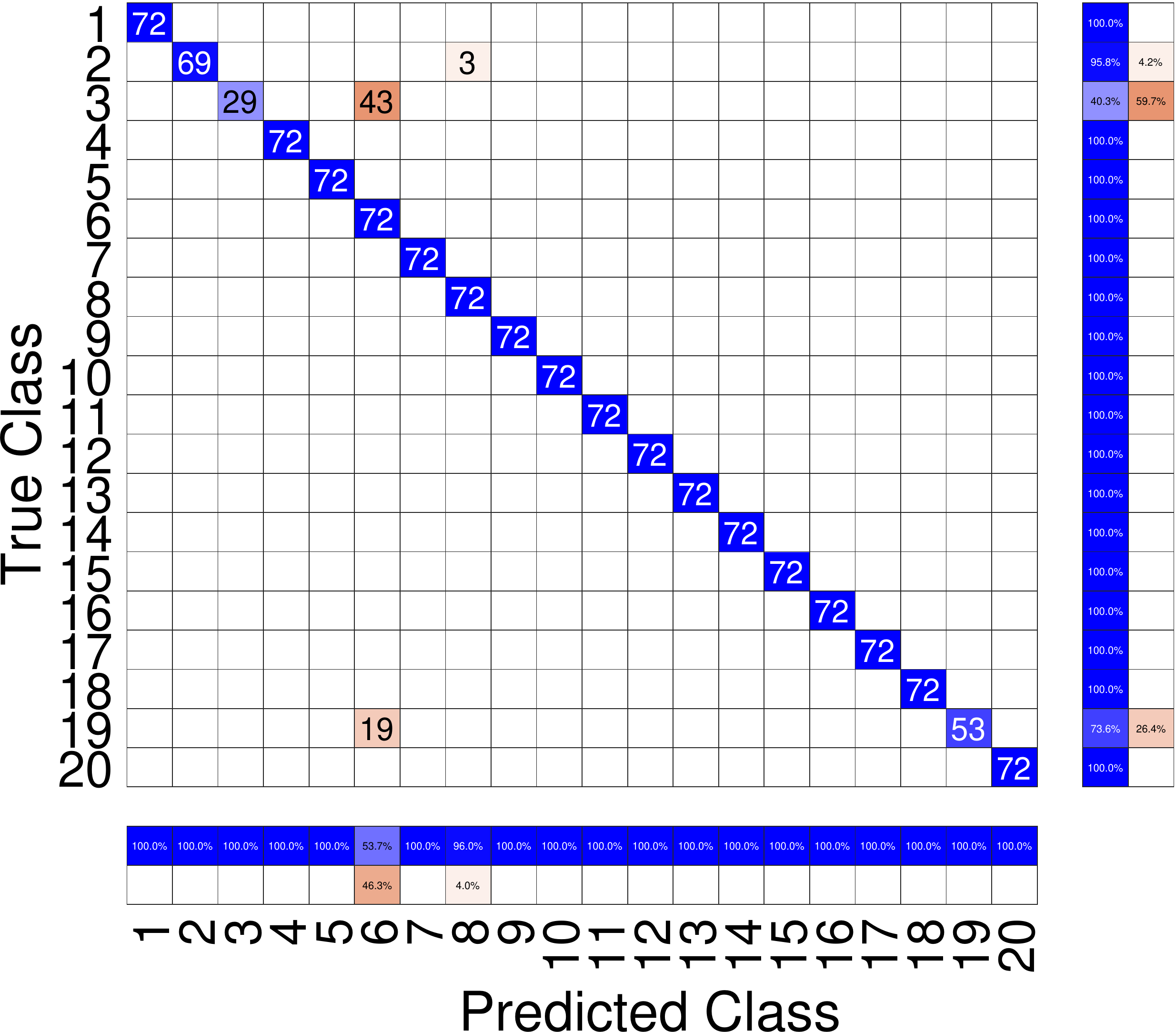}
		}
		\subfigure[DSC-Net-Nuclear-Norm]{
			\includegraphics [width=0.58\columnwidth]{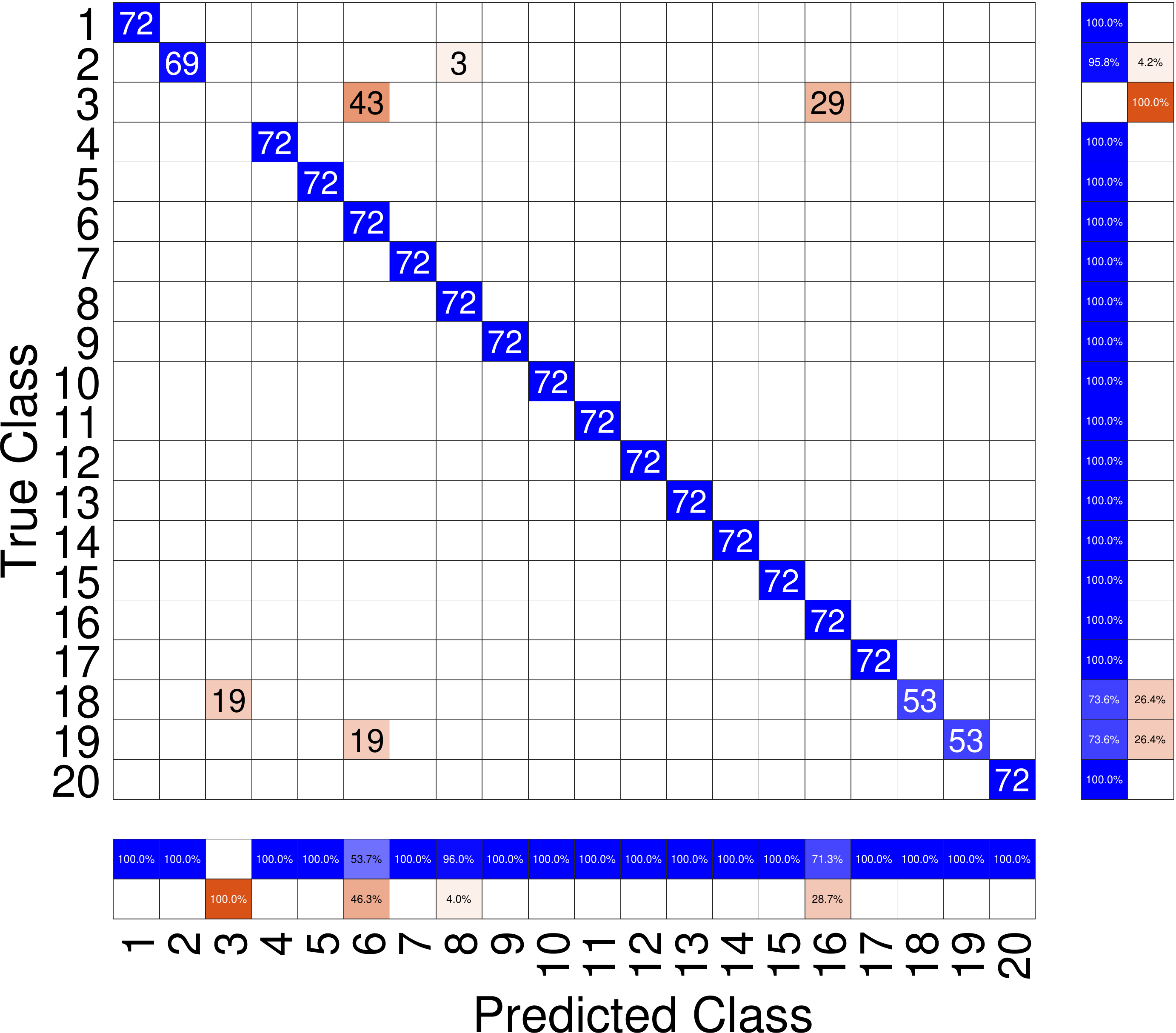}
		}
		\subfigure[DSLSP-L1]{
			\includegraphics [width=0.58\columnwidth]{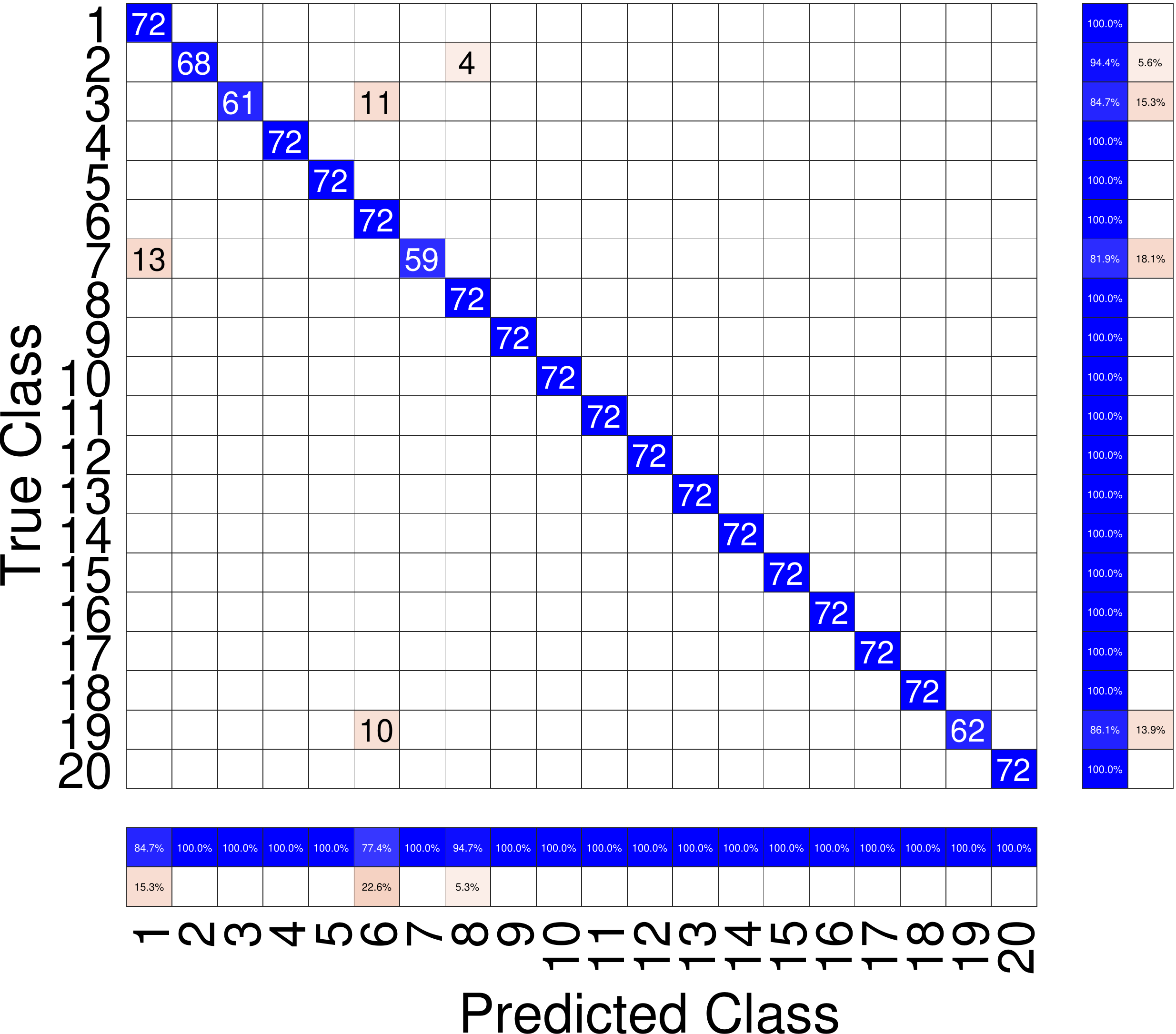}
		}
		\subfigure[DSLSP-L2]{
			\includegraphics [width=0.58\columnwidth]{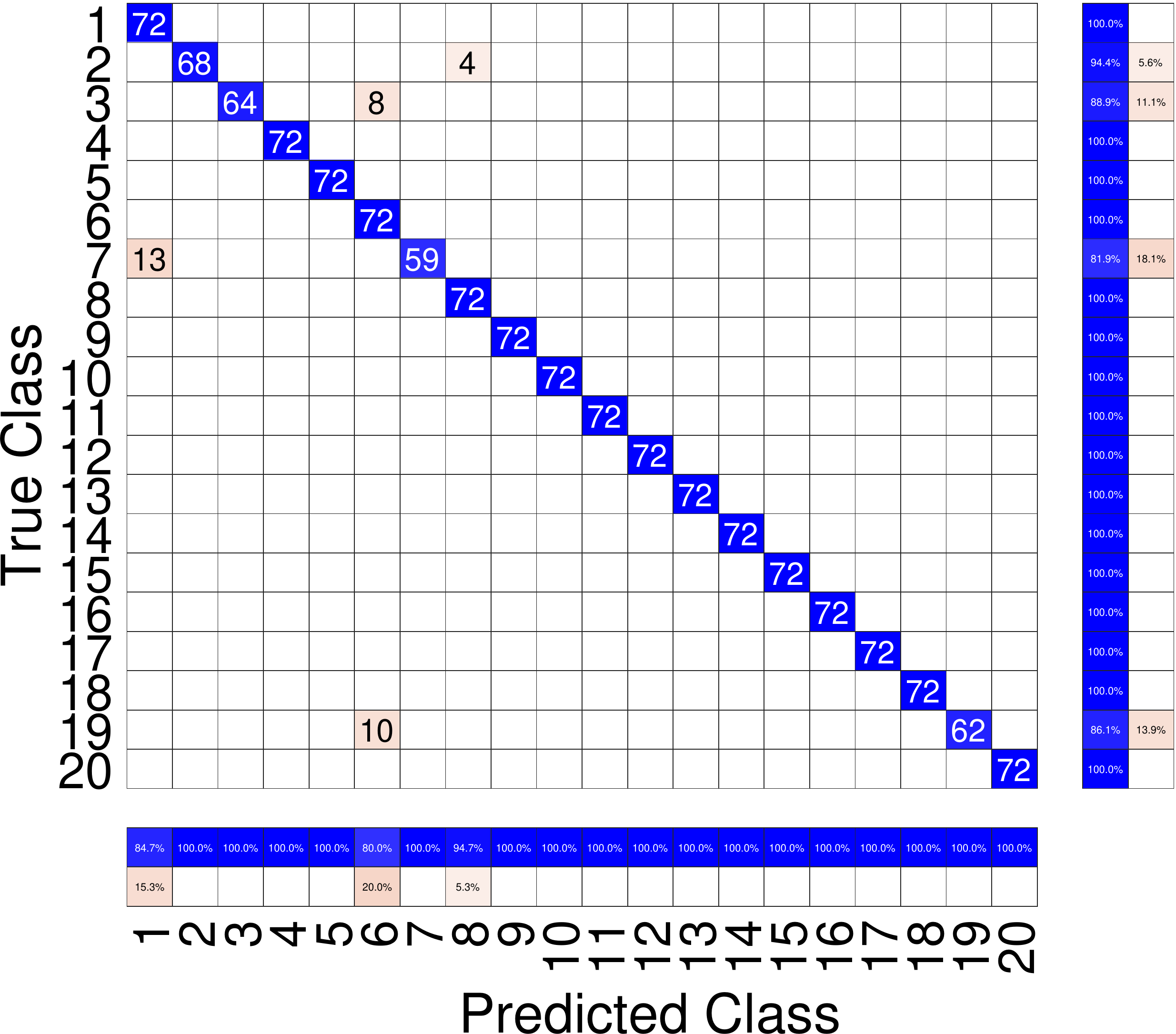}
		}
		\subfigure[Our]{
			\includegraphics [width=0.58\columnwidth]{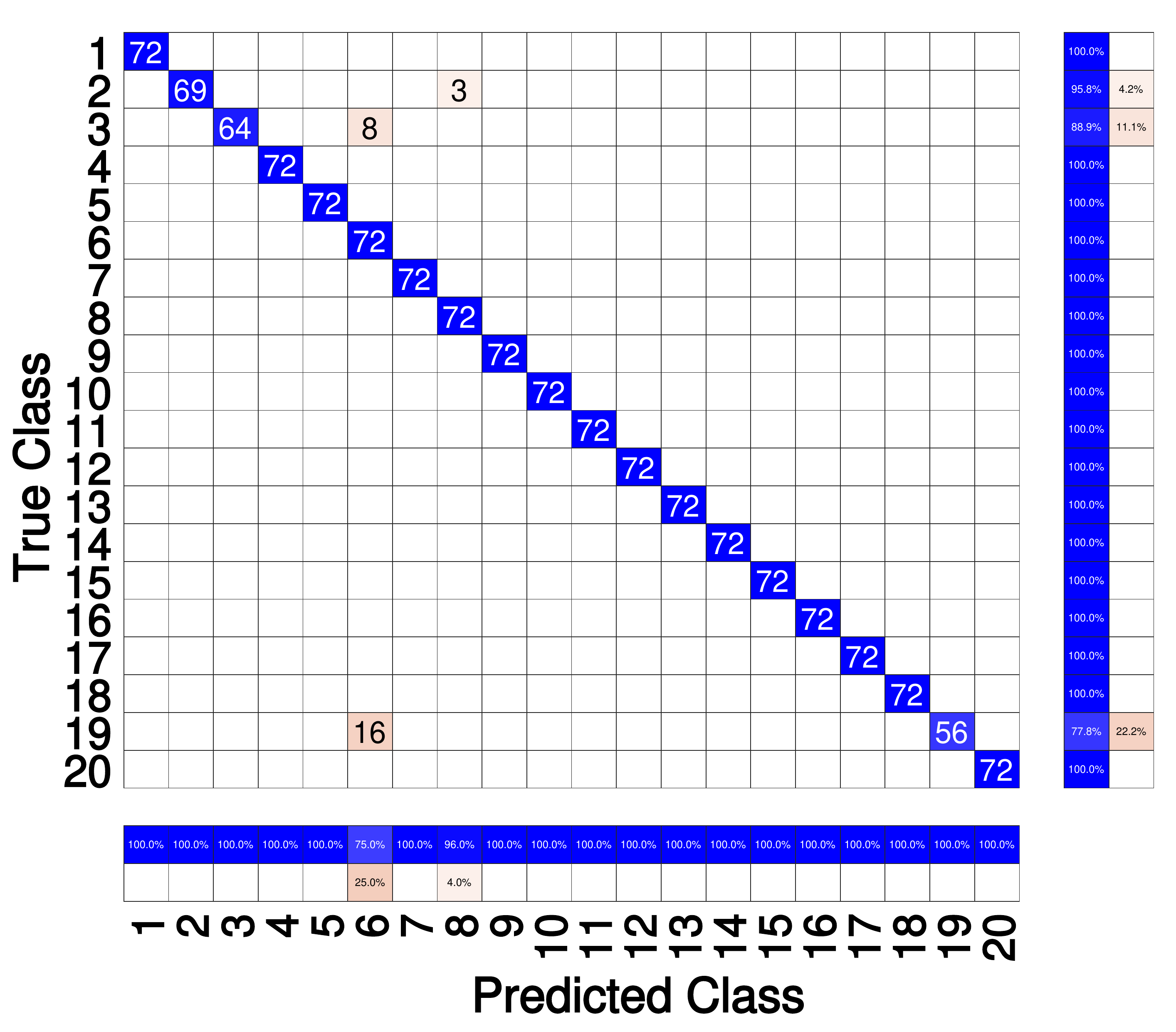}
		}	
		\caption{Confusion matrices on COIL20 by different methods, where the right row summary denotes the recall, and the below column summary denotes the precision.}
		\label{fig:cfm}
	\end{figure*}

	\subsection{Implementation Details}
	{
	For fair comparisons, we followed the same settings and most of the implementation choices as \cite{ji2017deep}. 
	First, we pre-train the deep auto-encoder without considering the self-expressiveness loss term, in which the deep auto-encoder can produce a suited representation. Then, in the fine-tuning stage, all the data are used to minimize the overall loss function with a gradient descent method. Specifically, we used ADAM \cite{kingma2014adam} to minimize the loss for all our experiments. 
	}
	The network was implemented with TensorFlow. Since ADAM may produce negative value during the gradient descent process, we performed $\max\left(\varepsilon,c_{i,j}\right)$ to maintain the feasibility of the logarithmic function (i.e., $\ln\left(c_{i,j}\right)$), where $\varepsilon$ (set as $1.0 \times 10^{-12}$ in our experiment) is a minimal positive value. In TensorFlow, the $\max\left(\varepsilon,c_{i,j}\right)$ can be achieved by the `tf.clip\_by\_value' operation. Besides, the neural network parameters were initialized by the initializer `he\_normal' \cite{he2015delving}. For the stacked convolutional layers, we set the kernel stride as `$2$' in both horizontal and vertical directions and used Rectified Linear Unit (ReLU) \cite{krizhevsky2012imagenet} as the activation function. The learning rate was set to $1.0\times10^{-4}$ in all experiments, and the whole data set was used as one batch input. The network structure settings of our method for each dataset are listed in Table \ref{tab:datasets and NN setting}. Once the network is trained, we can perform spectral clustering on the affinity matrix $\mathbf{C}$ to generate the clustering result. The code will be publicly available.
	
	\begin{figure*}
		\centering
		\subfigure[Ground Truth]{
			\includegraphics [width=0.58\columnwidth]{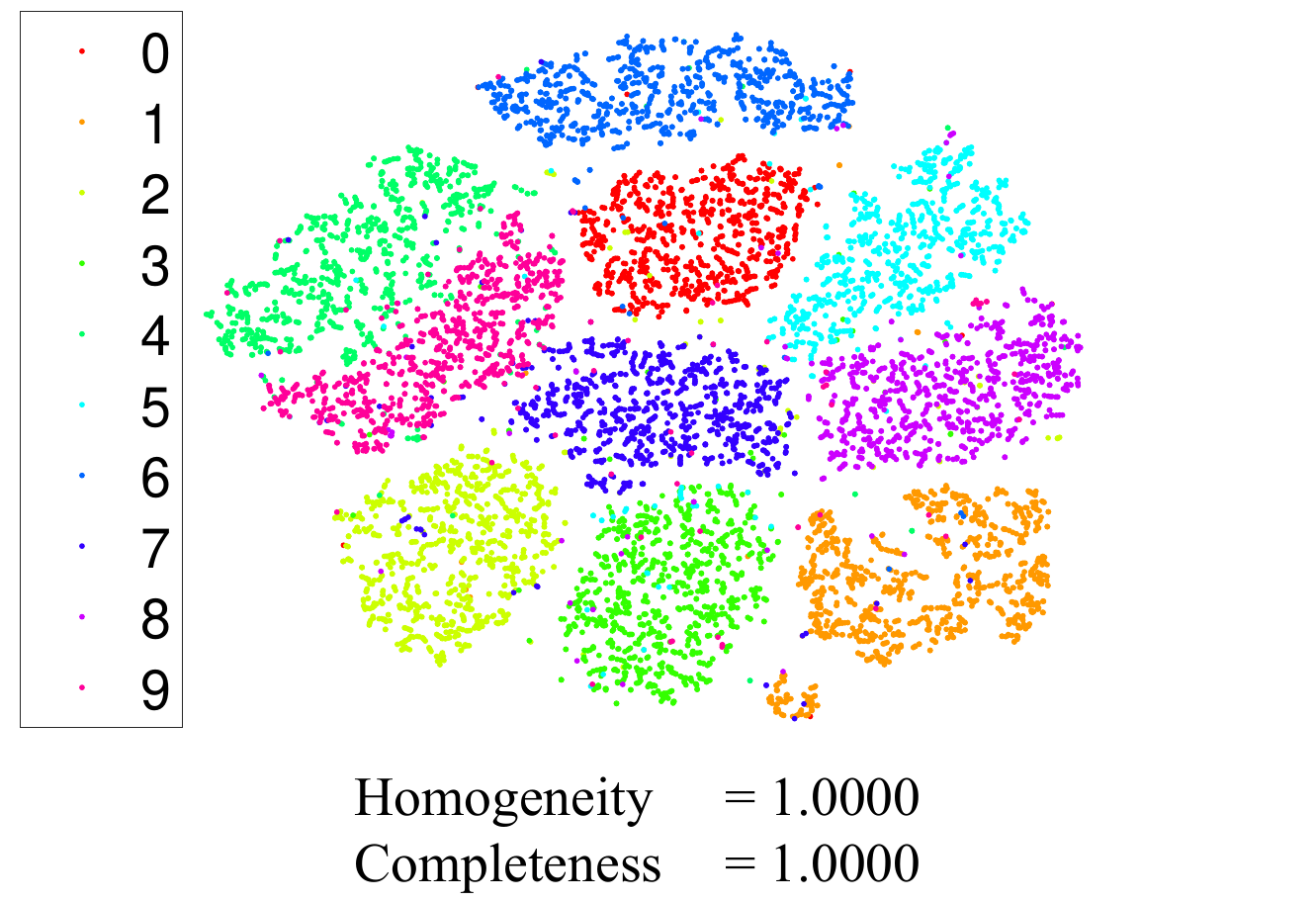}
		}
		\subfigure[DSC-Net-L1]{
			\includegraphics [width=0.58\columnwidth]{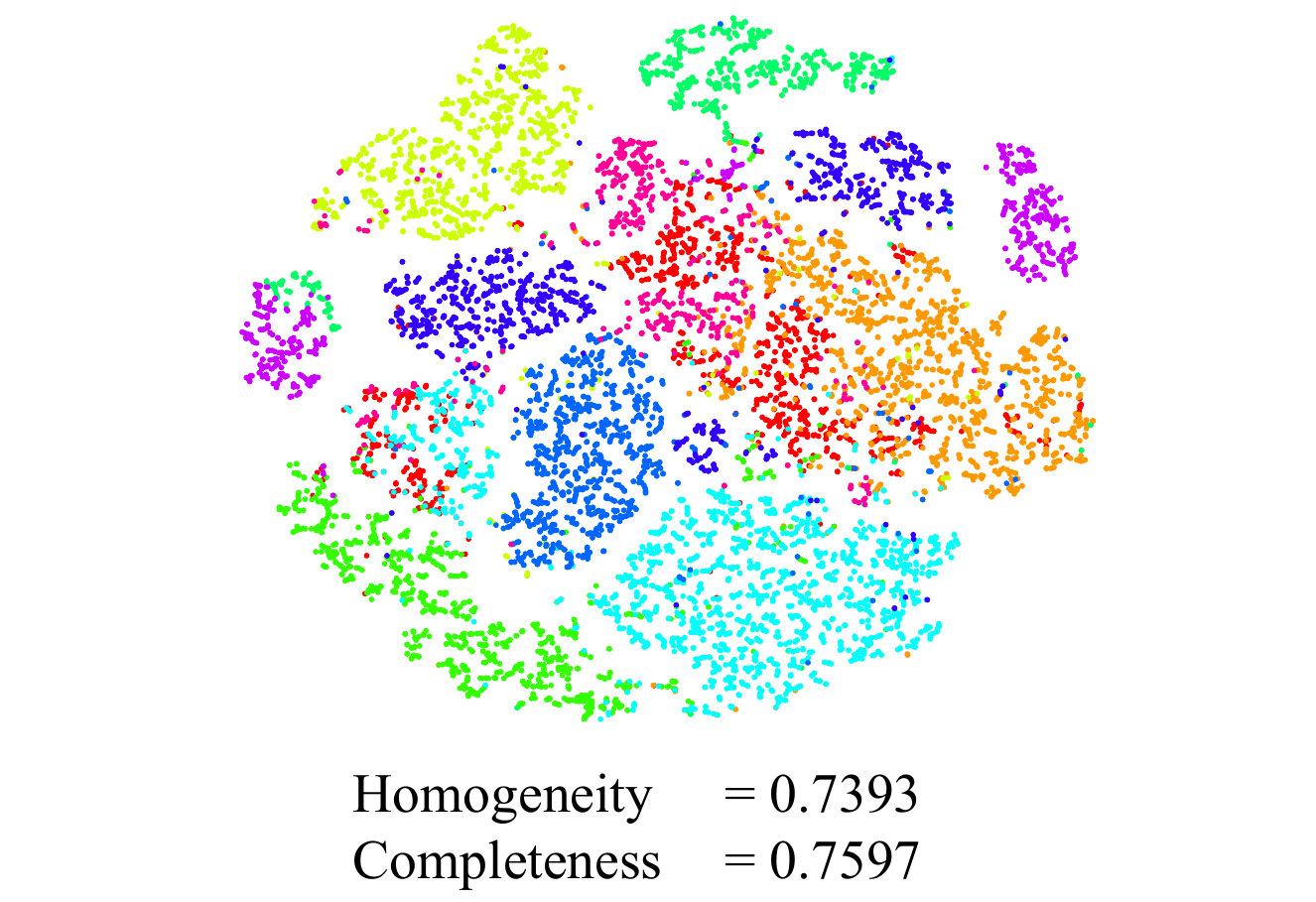}
		}
		\subfigure[DSC-Net-L2]{
			\includegraphics [width=0.58\columnwidth]{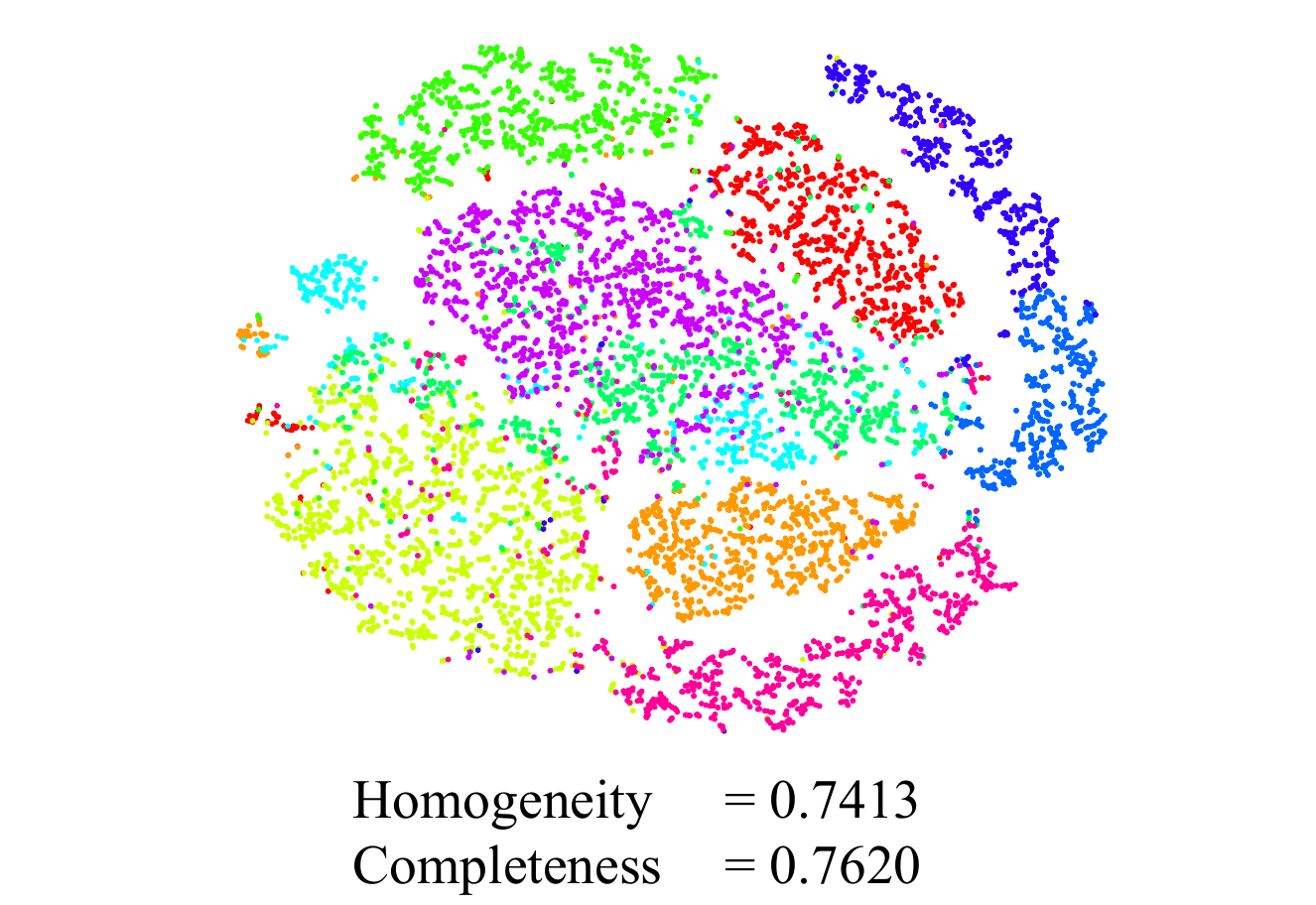}
		}
		\subfigure[DSLSP-L1]{
			\includegraphics [width=0.58\columnwidth]{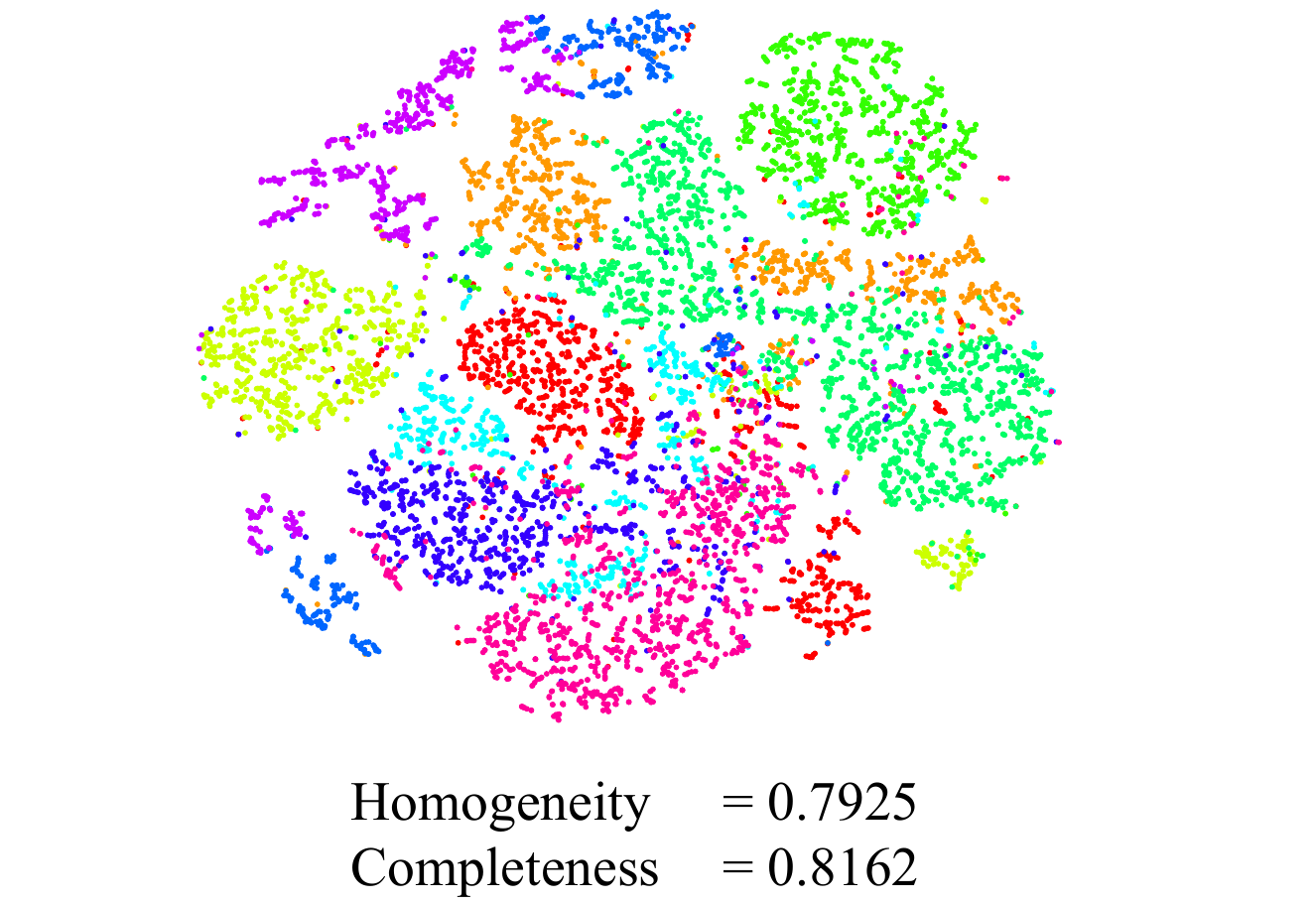}
		}
		\subfigure[DSLSP-L2]{
			\includegraphics [width=0.58\columnwidth]{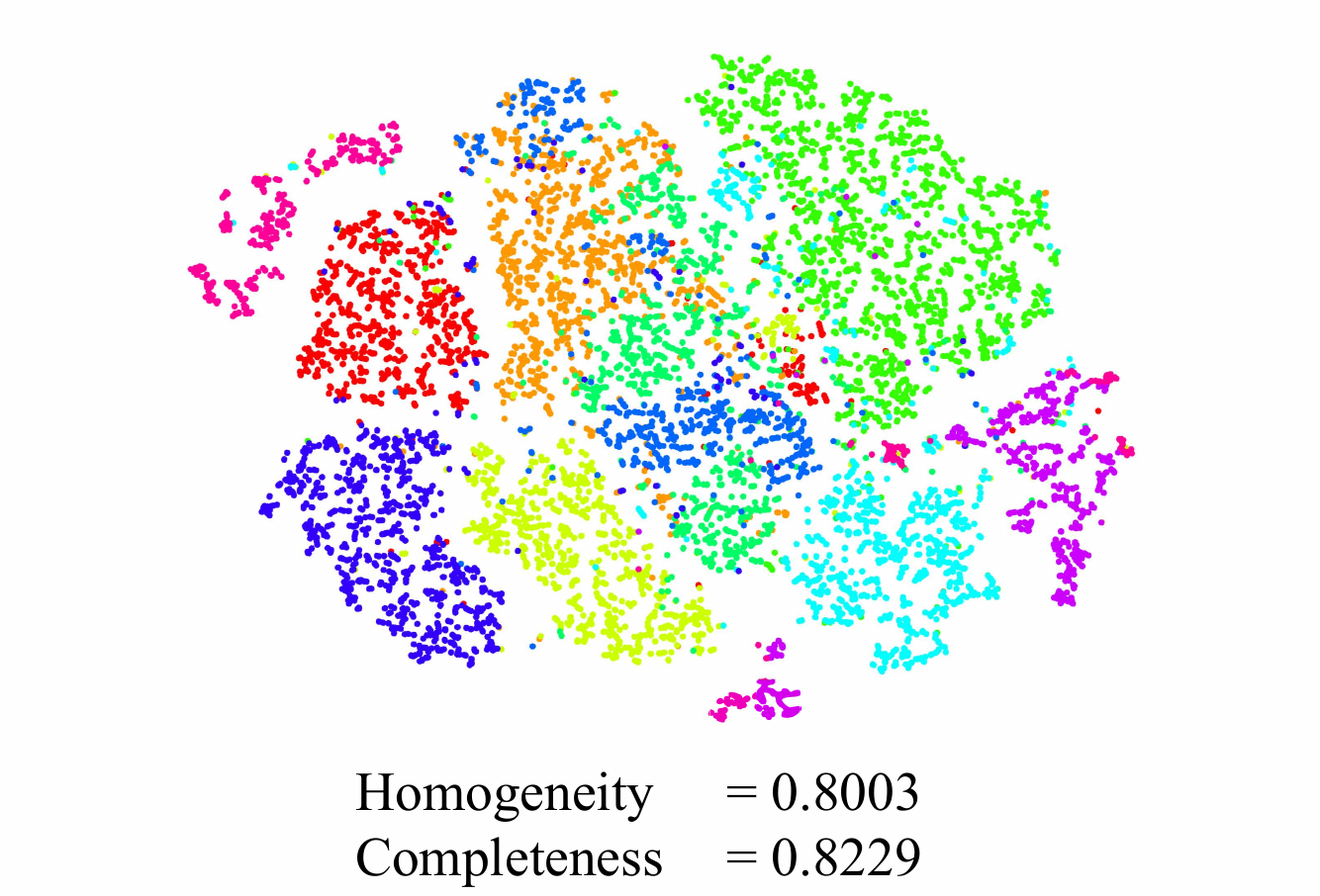}
		}
		\subfigure[Our]{
			\includegraphics [width=0.58\columnwidth]{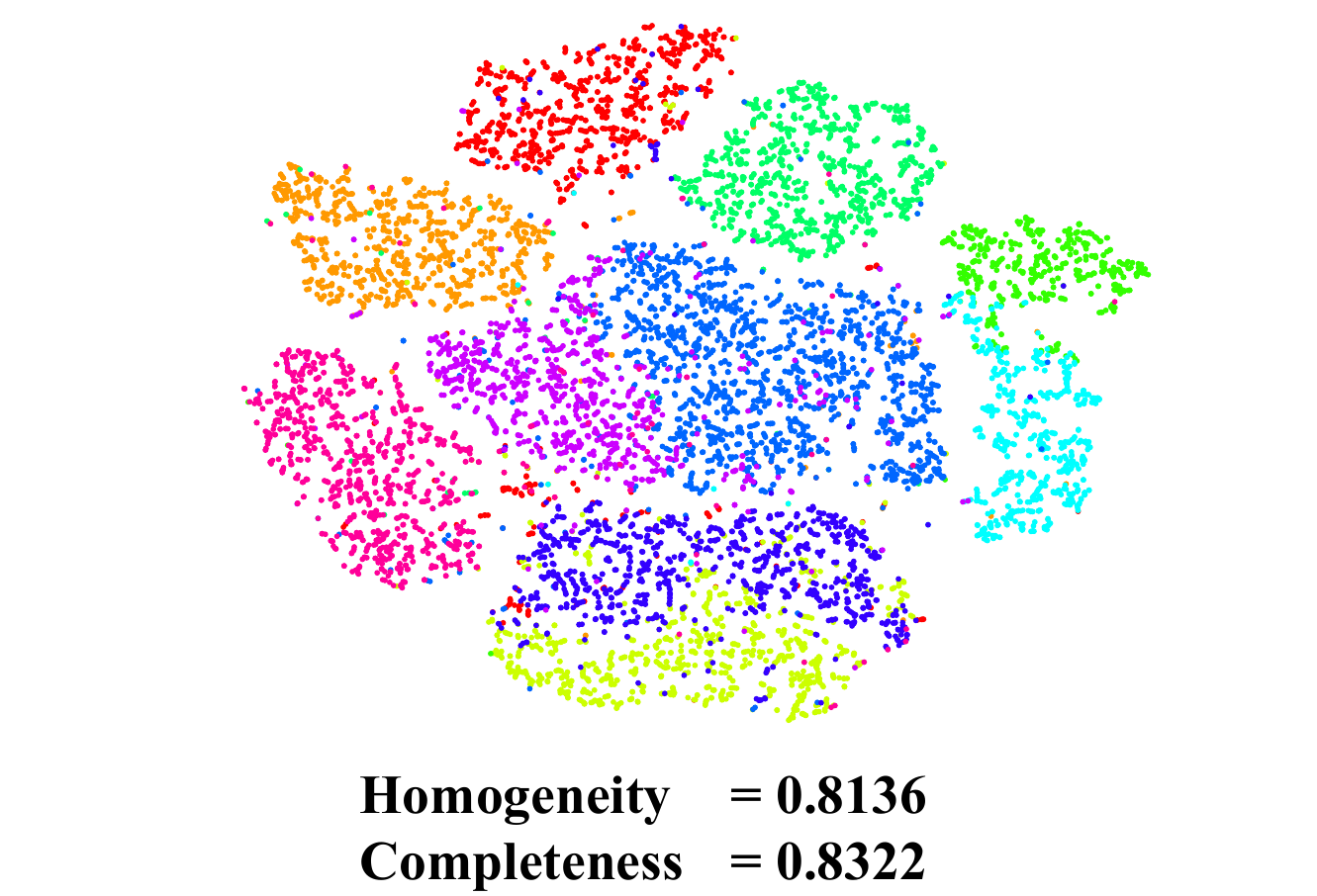}
		}
		\caption{Two-dimensional t-SNE plots of the encoder output representations on MNIST for (b) DSC-Net-L1, (c) DSC-Net-L2, (d) DSLSP-L1, (e) DSLSP-L2, and (f) Our. (a) denotes the t-SNE visualization of the ground-truth classes. Each color denotes a different class. Our representation promotes higher scores of homogeneity and completeness, which indicates better inter-class separability and intra-class compactness than the compared methods.}
		\label{fig:tsne}
	\end{figure*}
	\subsection{Clustering Results}
	As shown in Tables \ref{tab: ACC} and \ref{tab: NMI}, we can see that our methods (i.e., `Our w/o Net' and `Our') obtain the best clustering performance among all the methods in terms of both ACC and NMI. {Specifically, for MNIST, `Our' and `Our w/o Net' improve 7.67\% and 6.84\% over the DSC-Net-L1 and 7.5\% and 6.67\% over the DSC-Net-L2 in terms of ACC, and enhance 7.32\% and 5.33\% over the DSC-Net-L1 and 7.11\% AND 5.12\% over the DSC-Net-L2.} Note that the only difference between `Our w/o Net', DSC-Net-L1, and DSC-Net-L2 is the imposed regularizer on the affinity matrix. The performance advantages validate the proposed ME regularizer's effectiveness, i.e., the denser affinity matrix is conducive to obtain better spectral clustering results. {Furthermore, in terms of the comparisons with DSLSP-L1 and DSLSP-L2, it might be quite unfavorable and unfair for our method because DSLSP-L1 and DSLSP-L2 enhanced DSC-Net-L1 and DSC-Net-L2 by introducing an additional similarity preserving mechanism \cite{kang2020structure}. Notably, even without the additional similarity preserving mechanism, our method still keeps the superiority over DSLSP-L1 and DSLSP-L2.} 

	Moreover, in Figure \ref{fig:cfm}, we illustrated six confusion matrices on COIL20 by DSC-Net-L1, DSC-Net-L2, DSC-Net-Nuclear-Norm, DSLSP-L1, DSLSP-L2, and our MESC-Net. The row axis and column axis of the confusion matrices denote the predicted class labels and the ground-truth class labels, respectively; the right row summary and the below column summary denote recall and precision, respectively. 
	From Figure \ref{fig:cfm}, we can observe that our method wins in almost all classes on COIL20. Specifically, on `object 19' (convertible car), DSC-Net-L1 incorrectly clustered all samples. DSC-Net-L2 can not distinguish between `object 3' (model car) and `object 6' (toy car), `object 19'. Unlike them, our method increases the number and strength of samples' connections within the same subspace via entropy maximization, which correctly partitions to the utmost extent and produces a significant improvement (e.g., 56 correct samples on `object 19').

\begin{table*}[]
\caption{{Influence of ME regularization, decoupling framework, and pre-training on the ORL and COIL100 datasets. `$\checkmark$' and `$\times$' in each row denotes the usage of the ablation component and the corresponding reference, respectively. The best results are highlighted with \textbf{bold}.}}
\label{tab: AS}
\centering
\begin{tabular}{|c|l|l|l|l|l|l|l|l|l|}
\hline
\multicolumn{2}{|l|}{Index}                 & 1              & 2                & 3            & 4            & 5            & 6            & 7            & 8        \\ \hline
\multicolumn{2}{|l|}{ME regularization}     & $\checkmark$   & $\checkmark$     & $\checkmark$ & $\checkmark$ & $\times$     & $\times$     & $\times$     & $\times$ \\ \hline
\multicolumn{2}{|l|}{Decoupling framework}  & $\checkmark$   & $\checkmark$     & $\times$     & $\times$     & $\checkmark$ & $\checkmark$ & $\times$     & $\times$ \\ \hline
\multicolumn{2}{|l|}{Pre-training strategy} & $\checkmark$   & $\times$         & $\checkmark$ & $\times$     & $\checkmark$ & $\times$     & $\checkmark$ & $\times$ \\ \hline
\multirow{3}{*}{ORL}          
& ACC (\%)    & \textbf{90.25} & 86.00            & 89.75        & 73.00        & 86.50        & 82.75        & 86.00        & 68.75    \\
\cline{2-10} 
& NMI (\%)    & \textbf{93.59} & 92.42            & 93.23        & 84.65        & 91.94        & 89.76        & 90.34        & 81.53    \\
\cline{2-10} 
& Time (s)    & 110.55         & \textbf{13.27}   & 110.95       & 13.70        & 113.92       & 19.99        & 110.85       & 13.39    \\ \hline
\multirow{3}{*}{COIL100}    
& ACC (\%)    & \textbf{71.88} & 69.57            & 69.97        & 66.81        & 68.04        & 66.93        & 67.71        & 65.96    \\ 
\cline{2-10} 
 & NMI (\%)    & \textbf{90.76} & 90.31            & 89.86        & 89.24        & 89.43        & 88.87        & 89.08        & 88.44    \\ 
 \cline{2-10} 
& Time (s)    & 38873.41       & \textbf{2973.28} & 38785.97     & 3354.28      & 37840.35     & 3390.94      & 37778.14     & 3198.83  \\ \hline
\end{tabular}
\end{table*}

	\subsection{Ablation Study}
	{We conducted comprehensive ablation studies on the ORL and COIL100 datasets to evaluate the effectiveness and advantage of our method and the corresponding results are listed in Table \ref{tab: AS}.}

	\subsubsection{ME regularization} 
	{We started by examining the ME regularization, which underpins the key idea of our method, and reported the performance of \emph{Frobenius norm regularization} as a reference. For example, in term of the ORL dataset, by comparing index 1 with index 5, index 2 with index 6, index 3 with index 7, and index 4 with index 8, we can observe that the ME regularization produces a 3\% to 5\% performance improvement, which validates the effectiveness of the ME regularizer, i.e., focusing on the connectivity within each subspace is conducive to obtain better spectral clustering results.}

	\subsubsection{Decoupling framework} 
     We examined the advantage of the decoupling framework and reported the performance of \emph{DSC-Net framework} as a reference. By comparing index 1 with index 3, index 2 with index 4, index 5 with index 7, and index 6 with index 8, it is clear that the decoupling framework can help obtain better clustering performance. {Particularly, on the training process without pre-training, DSC-Net (i.e., index 6 vs. index 8) has a significant drop in clustering performance, while our decoupling framework (i.e., index 2 vs. index 4) can work stably in the overall performance.}

	\subsubsection{Pre-training}
	Pre-training is an important strategy of the model training for many previous works \cite{ji2017deep,chen2018subspace,kang2020structure}. We examined the performance contribution of pre-training in DSC-Net-L2 and our method w.r.t. the model performance. {We can observe that pre-training does help obtain good clustering performance at the cost of consuming more computation resources. Notably, the pre-training process requires more than 733.08\% training time (i.e., index 1 vs. index 2).}
	
    \subsection{Visual Comparison}
    Here we plot 2D t-distributed stochastic neighbor embedding (t-SNE) \cite{maaten2008visualizing} visualizations of the output representations of encoder for DSC-Net-L1, DSC-Net-L2, DSLSP-L1, DSLSP-L2 and our method on MNIST in Figure \ref{fig:tsne}. We observe that the t-SNE embedding produced by our method can separate each class visually. Since there is no globally-accepted metric for the intra-class compactness and inter-class separability in the existing literature, we made use of the heuristics employed by the previous work \cite{choi2020role} to quantitatively measure the clustering performance, i.e., the \textbf{homogeneity} and \textbf{completeness} scores. Specifically, a clustering result satisfies \textbf{homogeneity} if all of its clusters contain only data points that are members of a single class, and a clustering result satisfies \textbf{completeness} if all the data points that are members of a given class are elements of the same cluster. These metrics quantitatively measure the quality of the embeddings with a higher score indicating a better representation. As shown in Figure \ref{fig:tsne}, our method achieves the highest values on both homogeneity and completeness, which suggests that our method produces the most discriminative representation compared with state-of-the-art methods. 


	\begin{figure}
		\centering
		\subfigure[COIL20]{
			\includegraphics [width=0.46\columnwidth]{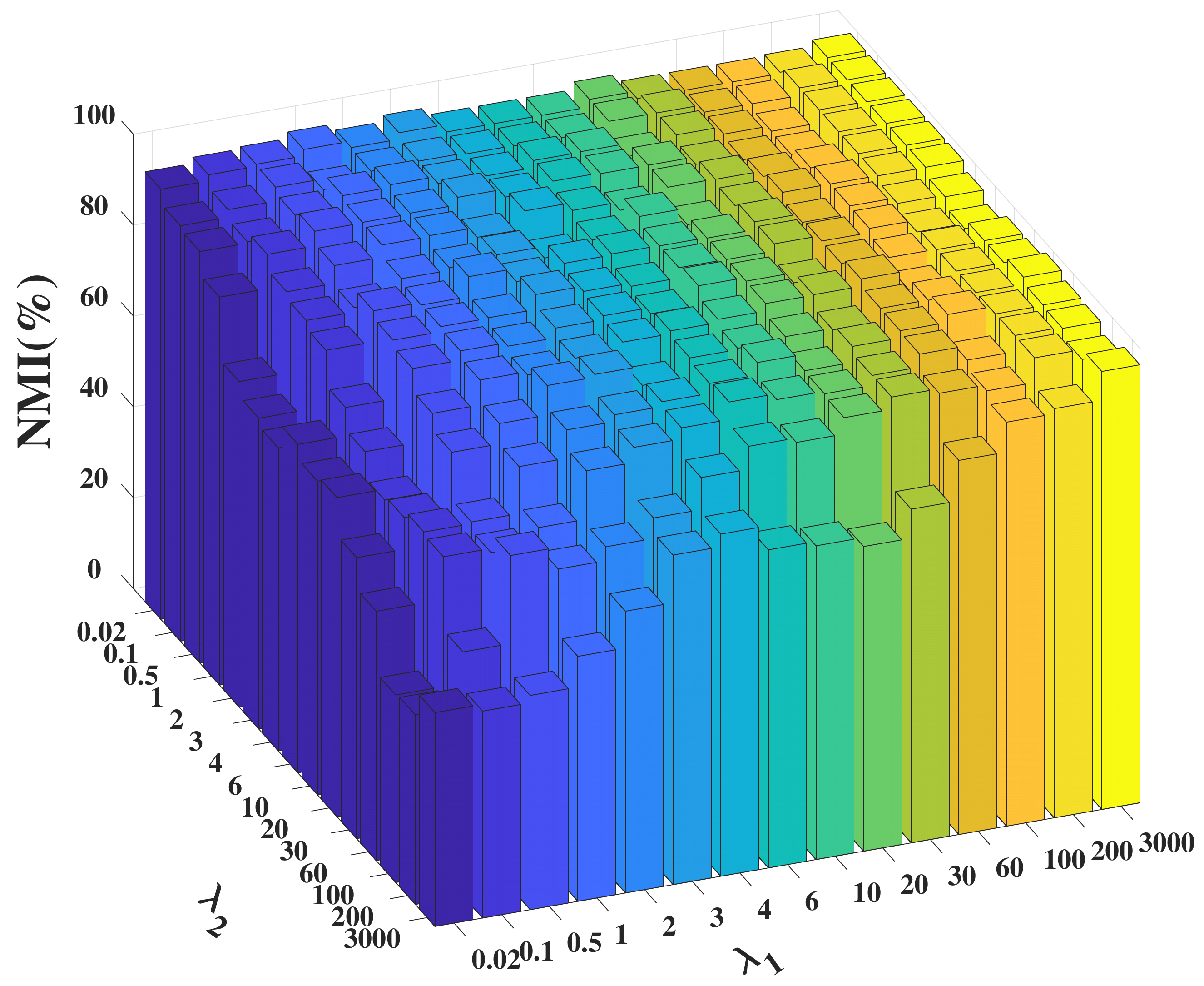}
		}
		\subfigure[ORL]{
			\includegraphics [width=0.46\columnwidth]{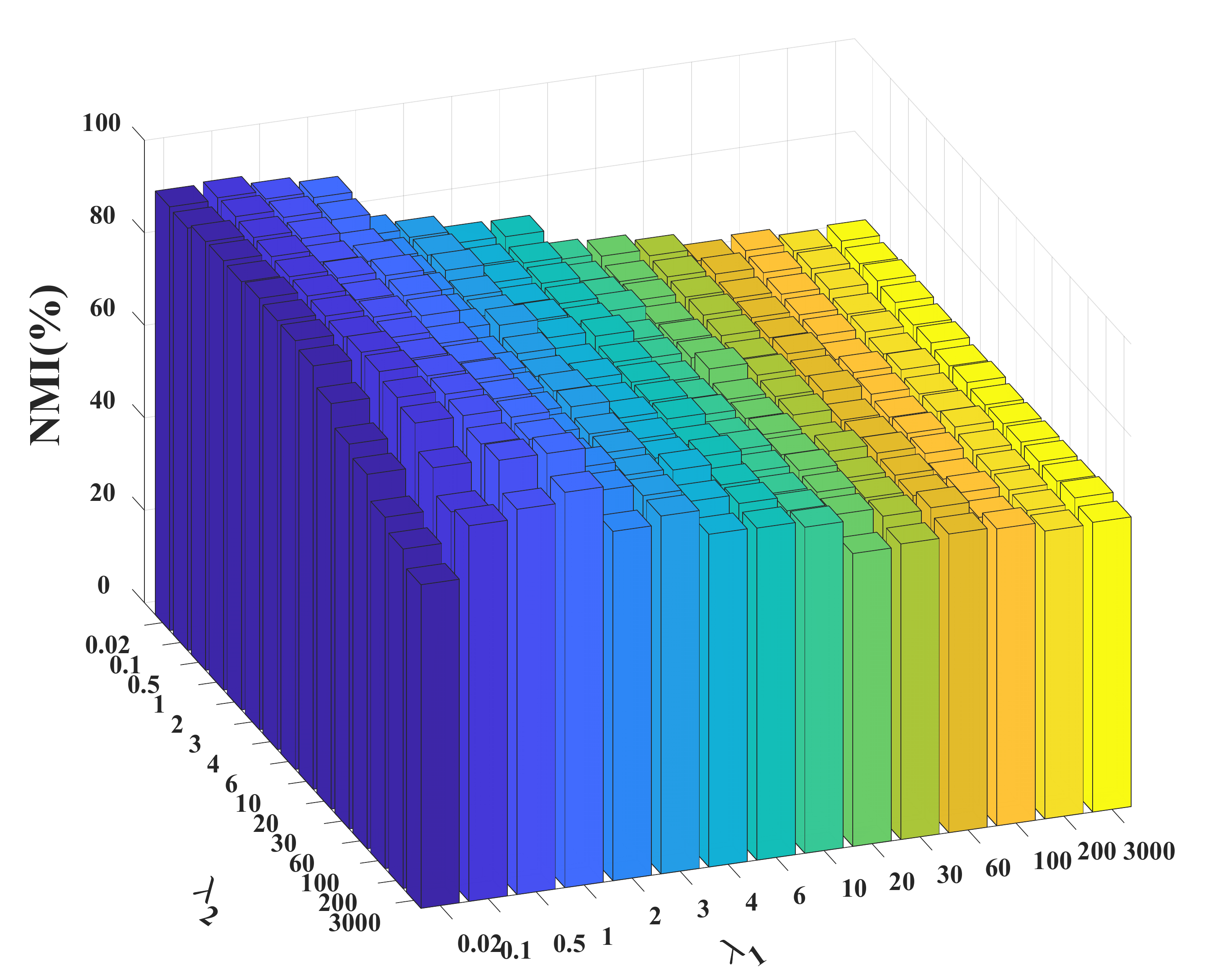}
		}	
		\caption{Sensitivity analysis of parameters $\lambda_1$ and $\lambda_2$ for our method on COIL20 and ORL.}
		\label{fig:para}
	\end{figure}
	
	\begin{figure}
		\centering
		\subfigure[ORL]{
			\includegraphics 
			[width=0.46\columnwidth]{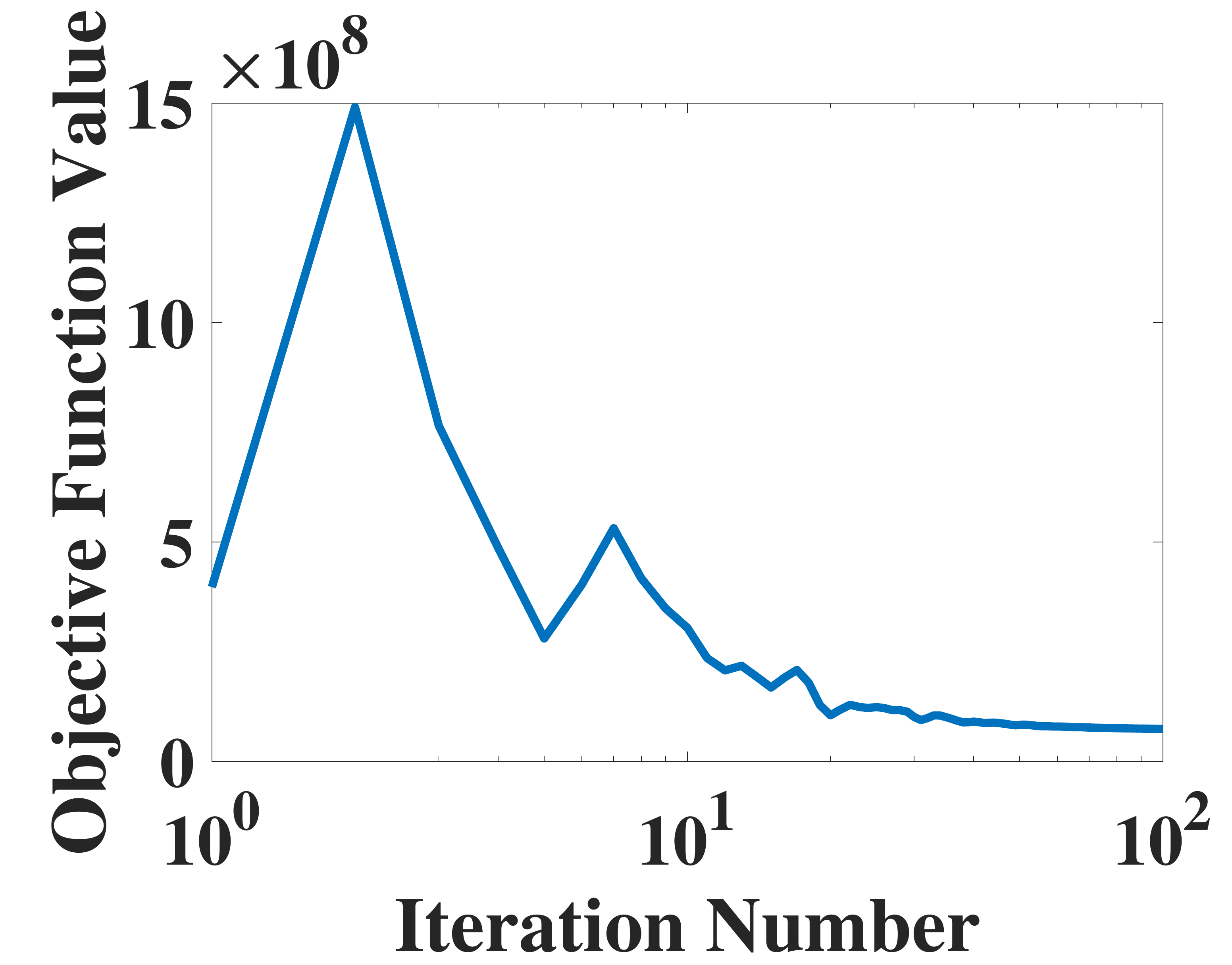}
		}
		\subfigure[COIL20]{
			\includegraphics 
			[width=0.46\columnwidth]{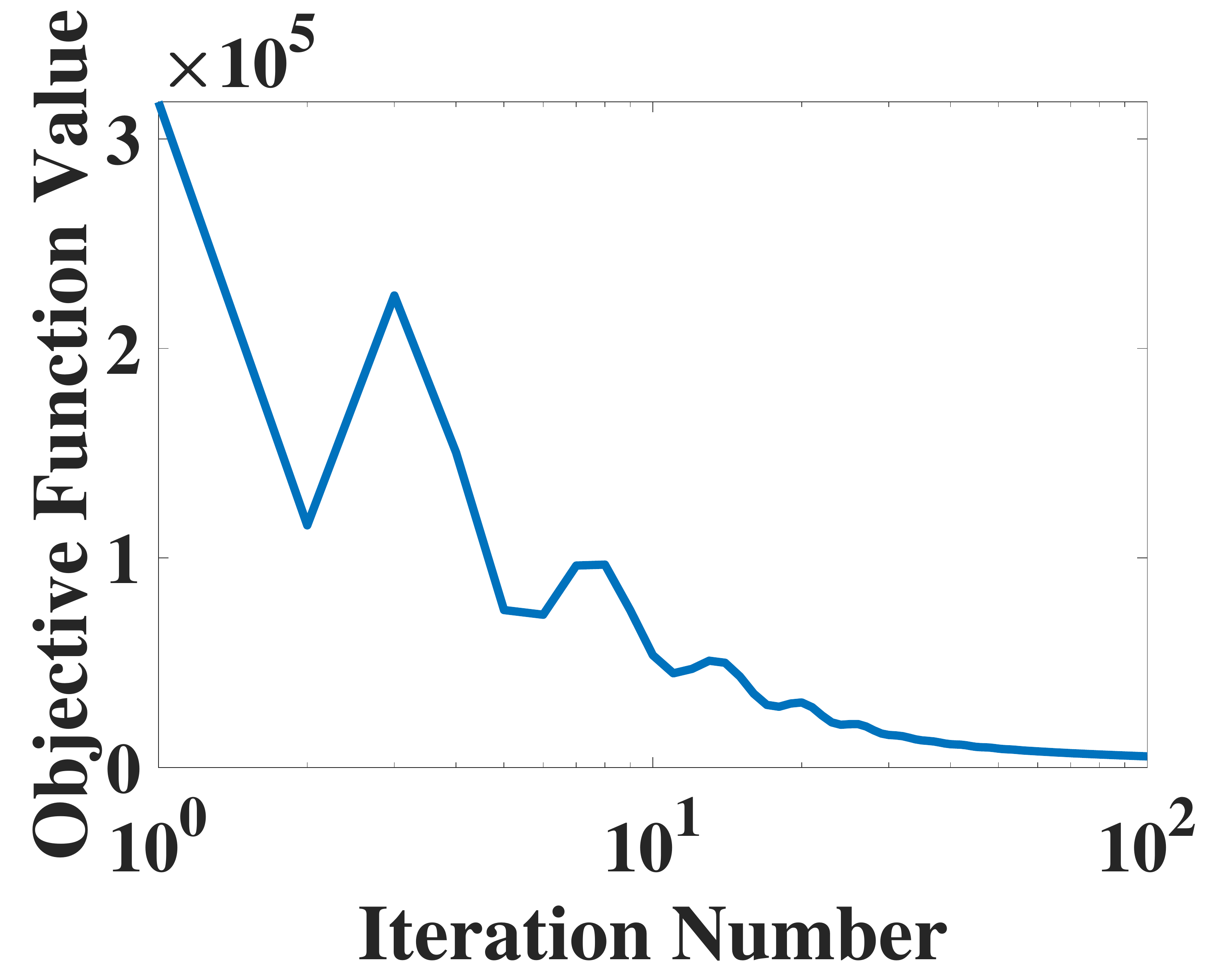}
		}
		\subfigure[EYaleB]{
			\includegraphics 
			[width=0.46\columnwidth]{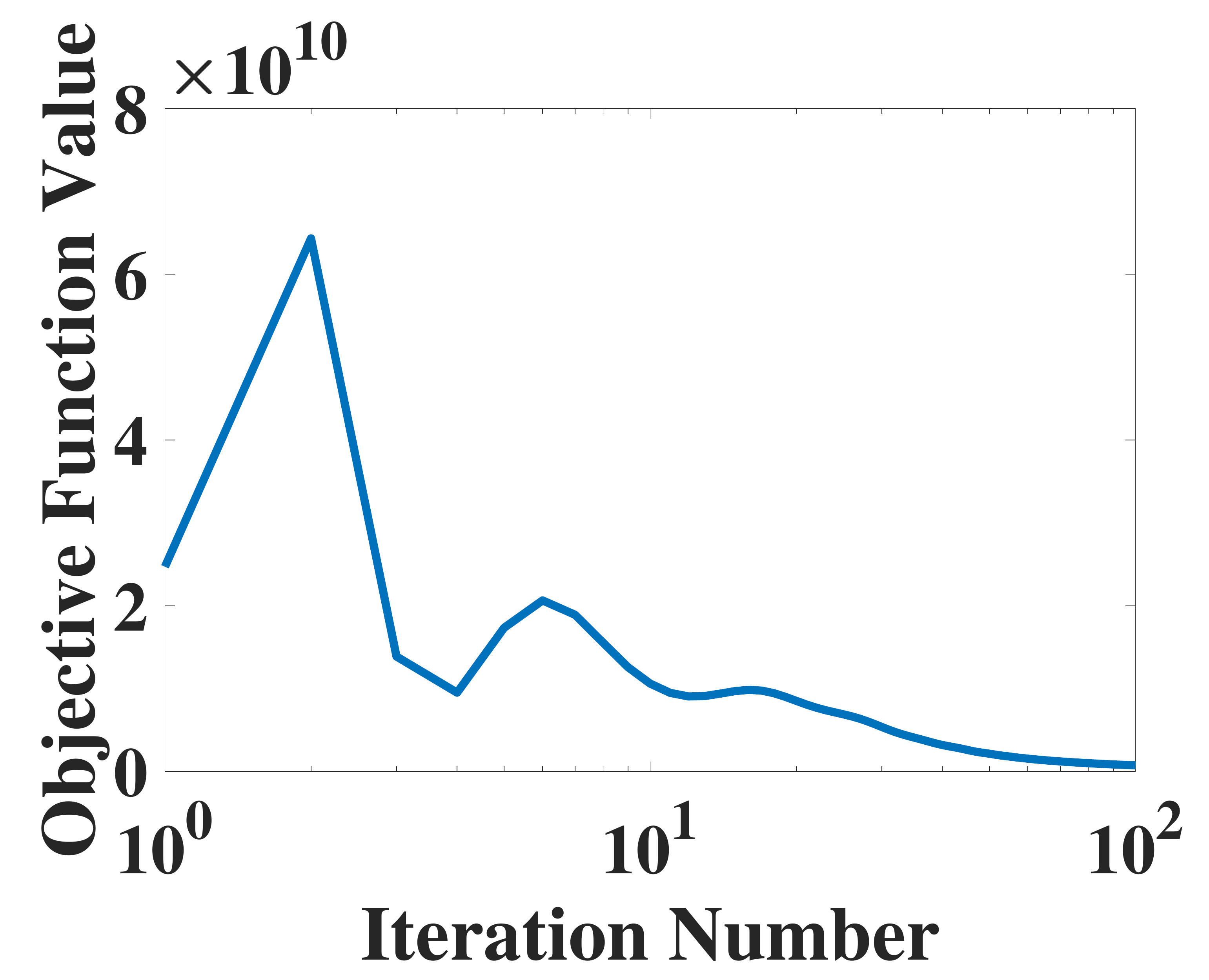}
		}
		\subfigure[COIL100]{
			\includegraphics 
			[width=0.46\columnwidth]{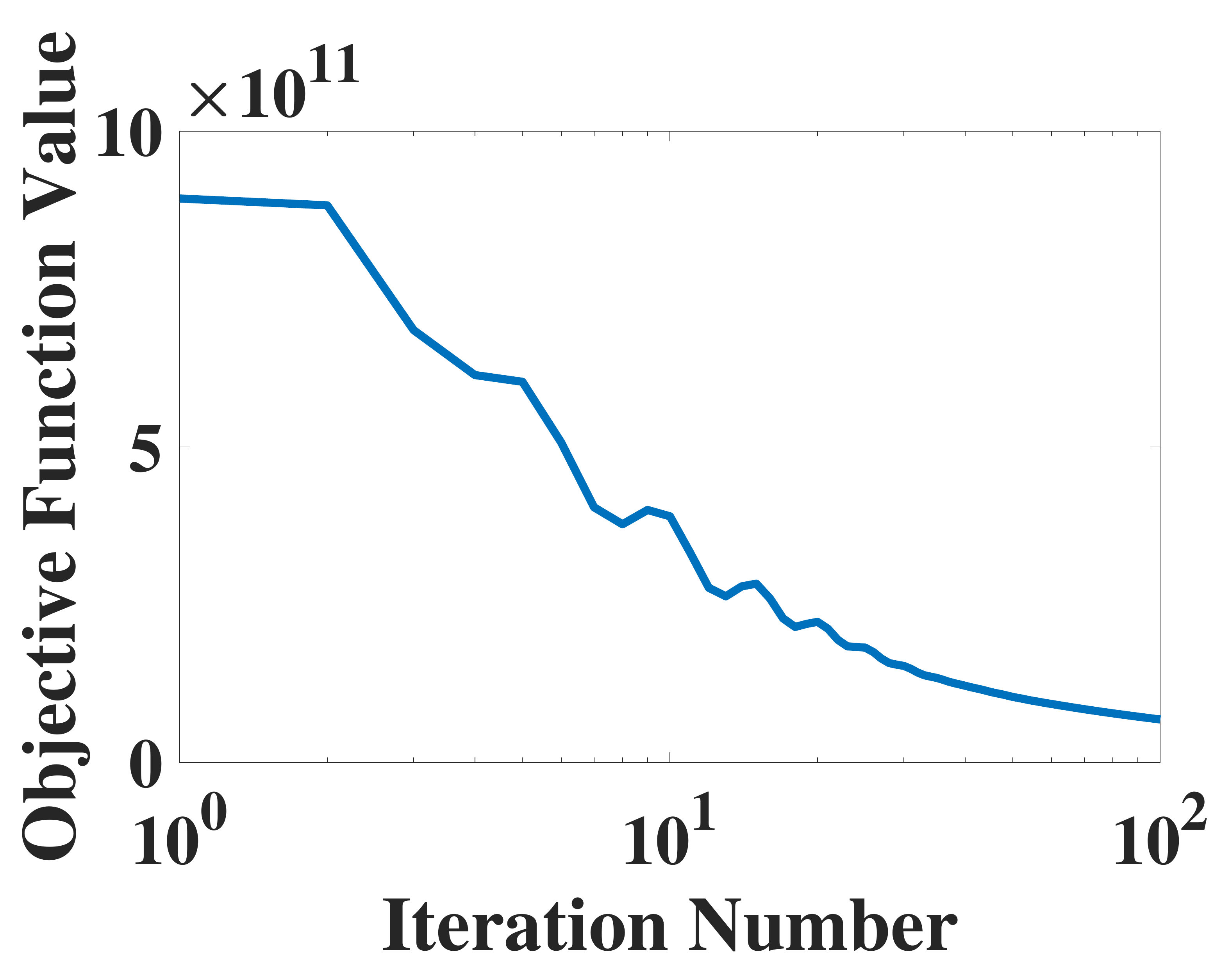}
		}
		\caption{The convergence curve of our method on four datasets.}
		\label{fig:conv}
	\end{figure}
	
	\subsection{Parameter Analysis}
	As shown in Eq. (\ref{eq:MESC-Net}), there are two hyper-parameters $\lambda_1$ and $\lambda_2$ in our objective function. In Figure \ref{fig:para}, we evaluated how they affect the clustering performance of MESC-Net, where we can observe that our algorithm is able to achieve the almost optimal clustering performance in a wide and common parameter range on different datasets, i.e., $\lambda_1\in\left[ 0.02, 2\right]$ and $\lambda_2\in\left[0.02, 30\right]$, indicating the robustness of our algorithm to the parameters and datasets. {More specifically, we observe that $\lambda_2$ should not be much larger than $\lambda_1$, reflecting the importance of the regularization term in constraining the affinity matrix. In a reasonable range of values, the regularization term is used to constrain the learning of the affinity matrix to obtain the appropriate appearances for subsequent spectral clustering.}

	\subsection{Convergence Analysis}
	Figure \ref{fig:conv} shows the convergence curves of our method on four datasets, where it can be observed that our method almost converges within 100 iterations on all the datasets. Specifically, in  Figure \ref{fig:conv} (a) and (c), the objective function values increase in the first several iterations. This phenomenon can be explained as the consequences of the initialization for different variables. In  Figure \ref{fig:conv} (d), there is an observation that the curve has a smooth change before the objective value starts to decrease sharply. This phenomenon is possible due to the influence of pre-training and fine-tuning strategies.

\section{Conclusion}

In this paper, we proposed MESC-Net, a novel deep learning-based subspace clustering method. Specifically, we employed the maximum entropy regularizer to strengthen the connectivity within each subspace, in which its elements corresponding to the same subspace are uniformly and densely distributed. We showed the visual illustrations of the learned affinity matrices to quantitatively and qualitatively validate the significant performance of MESC-Net. Besides, we theoretically prove that the learned affinity matrix satisfies the block-diagonal property, which is highly expected for clustering. Moreover, we explicitly decoupled the auto-encoder module and the self-expressiveness module, making the training process easier. Finally, we conducted extensive quantitative and qualitative experiments on commonly used benchmark datasets to demonstrate the superiority of MESC-Net over state-of-the-art methods. We also provided comprehensive ablation studies to validate the effectiveness and advantage of our network. In the future, we will also investigate the potential of our method in semi-supervised clustering \cite{wang2018unified} and classification \cite{shu2016image}.

\balance
\bibliographystyle{IEEEtran}
\bibliography{MESC-Net}

\end{document}